\documentclass[10pt,letterpaper]{article}
\usepackage[margin=1in]{geometry}
\usepackage[table]{xcolor}
\usepackage{graphicx}
\usepackage{times}
\usepackage{soul}
\usepackage{url}
\usepackage[hidelinks]{hyperref}
\usepackage[utf8]{inputenc}
\usepackage[small]{caption}
\usepackage{amsmath}

\usepackage{amsthm}
\usepackage{booktabs}
\usepackage{algorithm}
\usepackage{algorithmic}
\usepackage{xspace}
\usepackage{pifont}
\usepackage{xcolor}
\usepackage{multirow, bigstrut}
\usepackage{caption}
\usepackage{subcaption}
\usepackage{array}
\usepackage{enumitem}
\usepackage{verbatim}
\usepackage{bm}
\usepackage[labelfont=bf]{caption}
\usepackage{titlesec}

\newcommand{\ourtriplet}{\textsc{2STG}\xspace}
\newcommand{\ourfinetune}{\textsc{2STG+}\xspace}
\newcommand{\transfer}{\textsc{Tf}\xspace}
\newcommand{\lp}{\textsc{L2P}\xspace}
\newcommand{\ourtripletone}{\textsc{2STG} $=1$\xspace}
\newcommand{\ourtripletthree}{\textsc{2STG} $\leq3$\xspace}
\newcommand{\ourfinetuneone}{\textsc{2STG+} $=1$\xspace}
\newcommand{\ourfinetunethree}{\textsc{2STG+} $\leq3$\xspace}

\newcommand\red[1]{\textcolor{red}{#1}}

\newcolumntype{A}{>{\centering}p{0.12\textwidth}}
\newcolumntype{B}{>{\centering\arraybackslash}p{0.16\textwidth}}

\newcolumntype{C}{>{\centering}p{0.22\textwidth}}
\newcolumntype{D}{>{\centering\arraybackslash}p{0.22\textwidth}}
\begin{document}

\title{Two-Stage Training of Graph Neural Networks for Graph 
Classification}


\author{Manh Tuan Do\thanks{Kim Jaechul Graduate School of AI, KAIST, Seoul, South Korea manh.it97@kaist.ac.kr}         \and
        Noseong Park\thanks{Department of AI, Yonsei University, Seoul, South Korea, noseong@yonsei.ac.kr}		 \and
        Kijung Shin\thanks{Kim Jaechul Graduate School of AI \& School of Electrical Engineering, KAIST, Seoul, South Korea, kijungs@kaist.ac.kr}}

\date{}



\maketitle

\begin{abstract}
Graph neural networks (GNNs) have received massive attention in the field 
of machine learning on graphs. Inspired by the success of neural networks, 
a line of research has been conducted to train GNNs to deal with various 
tasks, such as node classification, graph classification, and link 
prediction. In this work, our task of interest is graph classification. 
Several GNN models have been proposed and shown great accuracy in this 
task. However, the question is whether usual training methods fully 
realize the capacity of the GNN models.
In this work, we propose a two-stage training framework based on triplet 
loss. In the first stage, GNN is trained to map each graph to a 
Euclidean-space vector so that graphs of the same class are close while 
those of different classes are mapped far apart. Once graphs are 
well-separated based on labels, a classifier is trained to distinguish 
between different classes. This method is generic in the sense that it is 
compatible with any GNN model. By adapting five GNN models to our method, 
we demonstrate the consistent improvement in accuracy and utilization of 
each GNN's allocated capacity over the original training method of 
each model up to $5.4\%$ points in 12 datasets.
\end{abstract}

\section{Introduction}
\label{sec:intro}

With the pervasiveness of graph-structured data, graph representation learning 
has become an important task. Its goal is to learn embeddings (i.e., vector 
representations) of nodes and/or (sub)graphs. These embeddings are used in 
various downstream tasks, such as node classification, link prediction, and 
graph classification. 

Metric learning is about
 learning distance between objects in a metric space. While it remains a 
 difficult task to properly define an effective metric measure directly based 
 on graph topology, a common approach is to map the graphs into vectors in the 
 Euclidean space and measure the distance between those vectors. In addition to 
 satisfying the basic properties of metrics, this mapping is also expected to 
 separate graphs of different classes to distinguishable clusters.

Graph neural networks (GNNs) have received a lot of attention in the graph mining 
literature. Despite the challenge of applying the message-passing mechanism of 
neural networks to the graph structure, GNNs 
have proved successful in dealing with graph learning problems, including 
node 
classification ~\cite{velickovic2018graph,kipf2016semi}, link 
prediction~\cite{schlichtkrull2018modeling} and graph 
classification~\cite{zhang2018end,dai2016discriminative,duvenaud2015convolutional}.
The common approach is to 
start from node features, allow information to flow among neighboring nodes 
and finalize the meaningful node embeddings. GNN models differ by the 
information-passing method and the objectives of the final embeddings.


Graph classification involves separating graph instances of different classes 
and predicting the label of an unknown graph. This task requires a graph 
representation vector 
distinctive enough to distinguish graphs of different classes. The 
subtlety is how to combine the node embeddings into an expressive graph representation vector,
and a number of approaches have been proposed.

Although GNNs are shown to achieve high  
accuracy of graph classification, we observe that, with usual end-to-end training methods, 
they cannot realize their full potential.
Thus, we propose \ourfinetune, a new training method with two stages.
The first stage is metric learning with triplet loss, and the second 
stage is training a classifier.
We observed that \ourfinetune significantly improves the accuracy of five 
different GNN models, compared to their original training methods.

Our training method is, to some extent, similar to~\cite{hlugzlplstrategies} and~\cite{lu2021learning}
in the sense that GNNs are pre-trained on a task before being used for 
graph 
classification. However, Hu et al.~\cite{hlugzlplstrategies} does transfer learning by 
pre-training GNNs on a different massive graph, either in chemistry or 
biology domain, with numerous tasks, on both node and graph levels. Lu et al.~\cite{lu2021learning} proposed the enhancement of pre-training steps  by simulating the fine-tuning process of down-stream tasks to help the model  adapt to future tasks. On the 
other hand, \ourfinetune pre-trains GNNs on the same training dataset with 
only one graph-level task as the first stage. As highlighted in 
Table~\ref{tab:compare}, \ourfinetune is faster without requiring pre-training on rich and massive
datasets, and it consistently achieves improved accuracy of more GNN models in more 
datasets than \cite{hlugzlplstrategies} and~\cite{lu2021learning}.

In short, the contributions of our paper are three-fold.
\begin{itemize}
\item {\bf Observation}: In the graph classification task, GNNs often fail to 
exhibit their full power. Using a proper training method, their 
expressiveness can be further utilized.
\item {\bf Method Design}: We propose a two-stage learning method with pre-training based 
on triplet loss. With this method, up to 5.4\% points in accuracy can be 
increased. Moreover, our method also utilizes the capacity of each GNN better by producing embeddings with higher intrinsic dimension and weaker correlation between embedding dimensions.
\item {\bf Extensive Experiments}: We conducted comprehensive experiments with 5 
different GNN models and 12 datasets to illustrate the consistent improvement in accuracy and capacity utilization by 
our two-stage training method. We also compare our 
method with two strong graph transfer-learning methods to highlight 
the competency of our method.
\end{itemize}

\begin{table}
	\centering
	\caption{Comparison of our method \ourfinetune with 
	\protect\cite{hlugzlplstrategies} and \protect\cite{lu2021learning}. While they all need a pre-training step 
	before graph classification, our method consistently improves 
	accuracy of several GNNs even in datasets of a far domain while 
	requiring shorter training time and no additional rich dataset.}
	\label{tab:compare}
	\vspace{-1mm}
	\begin{center}
		\scalebox{0.85}{
			\centering
			\begin{tabular}{l||C|C|D}
				\toprule
				& \ \ourfinetune (Proposed) \ & \ Transfer Learning \cite{hlugzlplstrategies} & \ L2P-GNN \cite{lu2021learning}
				\\ \cmidrule{1-4}
				\multirow{2}{*}{Accuracy improvement \ }  & 5 out of 5 GNNs & 3 out of 4 GNNs & 4 out of 4 GNNs\\ 
				 & All datasets & Only within domain & Only within domain
				\\
				\midrule     
				Required datasets & No additional set & Large ($\approx$ 400K 
				graphs) & Large ($\approx$ 400K 
				graphs)
				\\  
				\midrule   
				Total training time & Short ($\approx$ 1 hour) & Long ($\approx$ 1 day) & Long (several hours)\\
				\bottomrule
			\end{tabular}
		}
	\end{center}
\vspace{-0.5mm}
\end{table}

The rest of this paper is organized as follows.
In Section~\ref{sec:background}, we review some relevant studies. In 
Section~\ref{sec:method}, we define the problem and describe our method. The 
experiments and results are presented in Section~\ref{sec:experiments}. We 
conclude our work in Section~\ref{sec:conclusion}.

\section{Related work}
\label{sec:background}
\subsection{Graph neural networks}
Graph neural networks (GNNs) attempt to learn embeddings (i.e, vector representations) of nodes and/or graphs, 
utilizing the mechanisms of neural networks adapted to the topology of graphs. 
The core idea of GNNs is to allow messages to pass between neighbors so that 
the 
representation of each node can incorporate the information 
from its neighborhood and thus to enable the GNNs to indirectly learn the graph 
structures. Numerous novel architectures for GNNs have been proposed and 
tested, 
which differ by the information-passing mechanisms. Among the most recent 
architectures are graph convolutions~\cite{kipf2016semi},  attention 
mechanisms~\cite{velickovic2018graph}, and those inspired by convolutional 
neural networks~\cite{gao2019graph,niepert2016learning,defferrard2016convolutional}. The final 
embeddings obtained from GNNs can be utilized 
for various graph mining tasks, such as node 
classification~\cite{kipf2016semi}, link 
prediction~\cite{schlichtkrull2018modeling,hwang2022ahp}, graph 
classification~\cite{zhang2018end,dai2016discriminative}, and influence maximization~\cite{ko2020monstor}. In addition, GNNs have been applied to some real-time tasks in computer vision, including but not limited to object detection~\cite{shipoint}, learning human-object interactions~\cite{qilearning}, and region classification~\cite{xiescale}.

\subsection{Graph classification by GNNs}
In graph classification, GNNs are tasked with predicting the label of an unseen 
graph. While node embeddings can be updated within a graph, the elusive step 
here is how to combine them into a vector representation of the entire graph 
that can distinguish among different labels. Two of the most common approaches 
are global pooling~\cite{duvenaud2015convolutional} and hierarchical 
pooling~\cite{zhitao2018hierarchical,yao2019graph,lee2019self}. The simplest 
ways for 
global pooling are global entrywise mean and global entrywise max of the final 
node embeddings. In 
contrast, hierarchical pooling iteratively reduces the number of nodes either by merging 
similar nodes into supernodes~\cite{zhitao2018hierarchical,yao2019graph} or 
selecting the most significant nodes~\cite{lee2019self} until reaching a final 
supernode whose embedding is used to represent the whole 
graph. 

\subsection{Transfer learning for graphs}
While most existing methods attempt to train GNNs as an end-to-end 
classification system, some studies considered transfer learning in which the GNN 
is trained on a large dataset before being applied to the task of interest, often in 
a much smaller dataset. Hu et al.~\cite{hlugzlplstrategies} succeeded in 
improving 3 (out of 4 attempted) existing GNNs by transfer learning from other 
tasks. Rather than training a GNN to classify a dataset right away, the authors 
pre-trained that GNN on another massive dataset (up to 456K 
graphs);
then they added a classifier and trained the whole architecture on the graph 
classification task. Lu et al.~\cite{lu2021learning} also proposed a model consisting of two steps: pre-training and fine-tuning on downstream tasks. The key difference from ~\cite{hlugzlplstrategies} is that the model learns to adapt by simulating the fine-tuning process during the pre-training step. However, transfer learning for graph remains a major 
challenge, as Ching et al.~\cite{ching2018opportunities} and Wang et al.~\cite{wang2019data} pointed out, considerable domain knowledge is needed to design the appropriate 
pre-training procedure.

\subsection{Metric learning}
Metric learning aims to approximate a real-valued distance between two objects, and the most common objectives are contrastive loss and triplet loss. 
Some work has focused on metric learning on 
graphs~\cite{ktena2018metric,liu2019community,ling2020hierarchical}. Ktena et al.~\cite{ktena2018metric} and Liu et al.~\cite{liu2019community}
employ a Siamese network structure, in which a twin network sharing the same 
weights is applied on a pair of graphs, and the two output vectors acting as 
representation of the two graphs are passed through a distance measure.

In computer vision, Schroff et al.~\cite{schroff2015facenet} and Chechik et al.~\cite{chdchik2010large} learn metric on 
triplet 
of images, where two (anchor and positive) share the same label and  
one (negative) has a different label. The model aims to
minimize the distance between the anchor and the positive, while 
maximizing the distance between the anchor and the negative. This inspired our 
interest in learning graph metrics with a triplet loss.

\section{Proposed method}
\label{sec:method}

In this section, we first define our task of interest: graph classification. 
We then describe each component of our proposed training method of GNNs for 
graph classification.

\paragraph{Problem definition.\\}
We tackle the task of graph classification. Given 
$\mathcal{D}=\{(G_1,y_1),
(G_2,y_2),...\}$ where $y_i \in \mathcal{Y}$ is the 
class label of the graph $G_i \in \mathcal{G}$, the goal of graph 
classification is to learn a mapping $f : \mathcal{G} \to \mathcal{Y}$ that 
maps graphs to the set of class labels and predicts the class labels of unknown 
graphs. 

\paragraph{Outline of our method.\\}
Our method combines the advantages of both GNN and metric learning.
Specifically, to facilitate a better accuracy of the classifier, our method 
first maps input graphs into vectors in the 
Euclidean space such that their corresponding vectors are well-separated 
based on classes. Below, we first briefly introduce GNNs and a learning 
scheme on triplet loss. 
We then describe the two stages of our method: pre-training a GNN and 
training a classifier.

\subsection{Graph neural networks}
Various GNNs have been proposed and proven effective in 
approximating such a function $f$. Starting from a graph $G=(V,E)$ with node 
features $X_G = \{\boldsymbol{x_v} | v \in V \}$, GNNs obtain final embeddings 
of nodes $Z_G = \{\boldsymbol{z_v} | v \in V \}$ and a final embedding \boldsymbol{$h_G$} of the 
graph after $K$ layers of aggregation. Specifically, at each $(k+1)$-th layer, 
the embedding \boldsymbol{$h_v^{(k+1)}$} of each node $v$ incorporates the embeddings of itself and its neighboring nodes $N(v)$ at the $k$-th layer as follows:
\begin{equation*}
	\boldsymbol{h_v^{(k+1)}} = MERGE\Big(\boldsymbol{h_v^{(k)}}, AGGR\Big(\{\boldsymbol{h_u^{(k)}} | u \in 
	N(v)\}\Big) \Big)
\end{equation*}
The embedding $h_G$ of the graph $G$ is then obtained by pooling all node 
embeddings into a single vector as follows:
\begin{equation*}
	\boldsymbol{h_G} = POOL\Big( \{\boldsymbol{h_v^{(i)}} | v \in V; i=1,..,K\}\Big)
\end{equation*}
GNNs differ by how the incorporating function $MERGE$, the 
aggregating function $AGGR$, and the final pooling function $POOL$ are 
implemented.

\begin{figure}[t]
	\centering
	\includegraphics[width=0.8\linewidth]{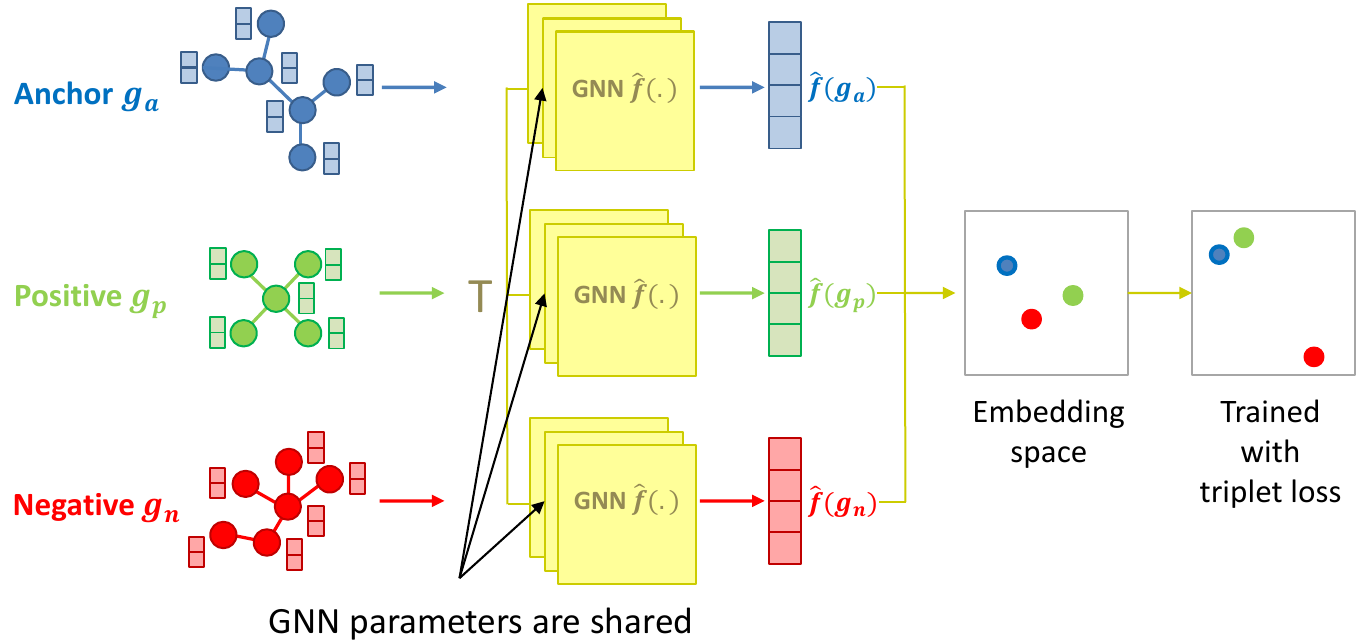} \\
	\vspace{-1.5mm}
	\caption{\label{fig:stage1}
		The first training stage for a GNN. $\hat{f}(.)$ is trained to 
		differentiate graphs of different classes via the triplet loss.
	}
\end{figure}

\begin{figure}[t]
	\centering
	\includegraphics[width=0.624\linewidth]{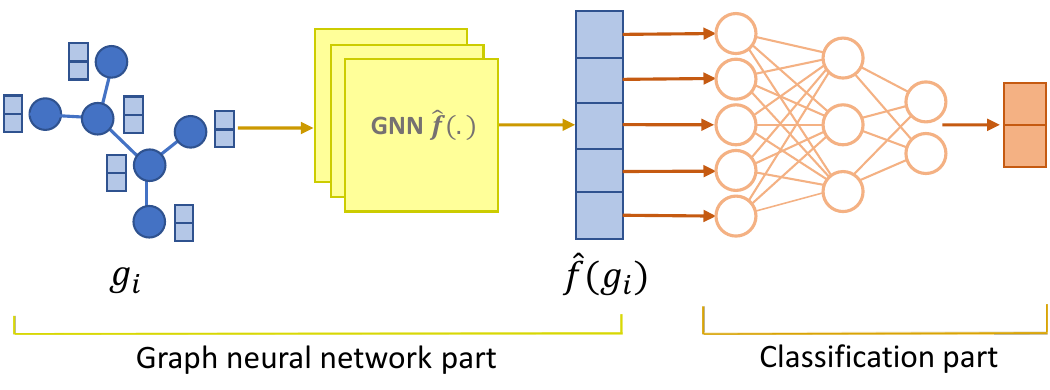} \\
	\vspace{-3mm}
	\caption{\label{fig:stage2}
		The second training stage for a classifier. After $\hat{f}(.)$ maps $g_i$ 
		to $\boldsymbol{\hat{f}(g_i)}$, a classifier is added to map 
		$\boldsymbol{\hat{f}(g_i)}$ to the 
		class probability prediction. At this step, either the classifier is 
		trained independently (\ourtriplet) or the whole architecture is 
		trained together (\ourfinetune).
	}
\end{figure}

\subsection{Metric learning based on triplet loss}
Triplet loss has been used for the learning of image 
similarity~\cite{schroff2015facenet,chdchik2010large}. The core idea 
is to enforce a 
margin between classes of samples. This results in embeddings of the same 
class mapped to a cluster distant apart from that of other classes. 
Specifically, given a mapping $\bar{f}$, we wish for a graph $g_a$ (anchor) 
to be closer to another graph 
$g_p$ (positive) of the same class than to a graph $g_n$ (negative) of another 
class by at least a margin $\alpha$, which is a hyperparameter:
$$\|\boldsymbol{\bar{f}(g_a)}-\boldsymbol{\bar{f}(g_p)}\|_2^2 + 
\alpha < 
\|\boldsymbol{\bar{f}(g_a)}-\boldsymbol{\bar{f}(g_n)}\|_2^2.$$

The triplet loss for the whole dataset becomes:
\begin{equation*}
\sum_{(a,p,n)}\max\left(\|\boldsymbol{\bar{f}(g_a)}-\boldsymbol{\bar{f}(g_p)}\|_2^2 - 
\|\boldsymbol{\bar{f}(g_a)}-\boldsymbol{\bar{f}(g_n)}\|_2^2 + 
\alpha, 0\right)
\end{equation*}
with the summation over all considered triplets.

Our two-stage method combines the power of both GNNs and the metric 
learning method, as described below.
\subsection{First training stage (pre-training a GNN)}
\label{subsec:firststage}
In the first training stage (depicted in Fig.~\ref{fig:stage1}), given a GNN architecture 
$\hat{f}(.)$, its weights are shared 
among a triplet network $T$, which consists of three identical GNN architectures having the same weights as $\hat{f}(.)$. The parameters of $T$ are trained on each triplet 
of graphs $(g_a,g_p,g_n)$ (anchor, positive, negative), in which the anchor and 
the positive graphs are of the same class while the negative graph is of 
another class. $T$ maps a triplet of graphs to a triplet of vectors in the Euclidean space: $T(g_a, g_p, g_n) = 
(\boldsymbol{\hat{f}(g_a)}, \boldsymbol{\hat{f}(g_p)}, \boldsymbol{\hat{f}(g_n)})$. Ideally, $\boldsymbol{\hat{f}(g_a)}$ and 
$\boldsymbol{\hat{f}(g_p)}$ should be close while $\boldsymbol{\hat{f}(g_n)}$ is far from them 
both. The triplet loss for $(g_a,g_p,g_n)$ is defined as: 
\begin{equation*}
	\max\left(\|\boldsymbol{\hat{f}(g_a)}-\boldsymbol{\hat{f}(g_p)}\|_2^2 - 
	\|\boldsymbol{\hat{f}(g_a)}-\boldsymbol{\hat{f}(g_n)}\|_2^2 + 
	\alpha,0\right)
\end{equation*}

\subsection{Second training stage (training a classifier)}
\label{subsec:secondstage}
In the second stage, a classifier is either trained 
independently, or added on top of the trained GNN and trained together on the 
graph classification task (see Fig.~\ref{fig:stage2}).

In summary, we propose two training methods for GNNs: \ourtriplet and 
\ourfinetune, both consist of two stages.
\begin{itemize}
	\item \textbf{\ourtriplet} (Pre-training Setting): In the first stage, the 
	GNN maps each triplet of graphs to a triplet of 
	Euclidean-space vectors, and the GNN is trained on triplet loss. In the 
	second stage, a classifier is trained independently to classify the graph 
	embeddings.
	\item \textbf{\ourfinetune} (Fine-tuning setting): It has the same structure as \ourtriplet 
	except that in the second stage, the classifier is plugged on top of 
	the trained GNN, and then the whole architecture is trained together in an end-to-end 
	manner.
\end{itemize}

Note that our methods are compatible to any GNN model that maps each 
graph to a representation vector. As shown in the next section, when applied to 
this method, each GNN model outperformed itself in the original setting.

\section{Experiments}
\label{sec:experiments}
In this section, we describe the details of our experiments.

\subsection{Experimental settings}
\label{sec:experiments:settings}

\subsubsection{GNN architectures}

In order to demonstrate that our two-stage training method helps realize a 
better 
performance of GNNs, for each GNN architecture, we compared the accuracy 
obtained in 
the original setting versus that from our method. The GNN 
architectures we considered in this work are:
\begin{itemize}
	\item \textsc{GraphSage}~\cite{hamiltoninductive}: This is often used as a 
	strong baseline in 
	graph classification. After obtaining node embeddings, global mean/max 
	pooling is applied to combine all node embeddings into one graph embedding.
	\item \textsc{GAT}~\cite{velickovic2018graph}: Instead of uniformly passing 
	neighbor information into 
	a node embedding, Veli{\v{c}}kovi{\'{c}} et al.~\cite{velickovic2018graph} uses an attention 
	mechanism for the 
	importance of each neighbor node.
	\item \textsc{DiffPool}~\cite{zhitao2018hierarchical}: While using the same 
	aggregation mechanism as in~\cite{hamiltoninductive}, 
	Ying et al.~\cite{zhitao2018hierarchical} proposes a 
	hierarchical approach to pool the node embeddings. 
	Rather than a ``flat-pooling" step at the end, \textsc{DiffPool} repeatedly merges 
	nodes into ``supernodes" until there is only one supernode whose embedding 
	is 
	treated as the graph embedding.
	\item \textsc{EigenGCN}~\cite{yao2019graph}: Attempting to implement 
	hierarchical 
	pooling like Ying et al.~\cite{zhitao2018hierarchical}, Ma et al.~\cite{yao2019graph} 
	formulates a different way to combine nodes and their respective embeddings 
	making use of the eigenvectors of the Laplacian matrix.
	\item \textsc{SAGPool}~\cite{lee2019self}: Hierarchical graph pooling 
	employing 
	self-attention mechanisms.
\end{itemize} 
In previous studies, these models were trained end-to-end, mapping each graph 
to a prediction of class probabilities.
To further illustrate the competency of our method, we also compared it with a 
transfer-learning method \cite{hlugzlplstrategies}. 

\subsubsection{Datasets}
We tested our training method using 12 datasets.
Some statistics of the datasets are summarized in 
Table.~\ref{tab:graph_stats} and~\ref{tab:graph_stats2}.
\begin{table*}[t]
	\caption{Some statistics of the benchmark datasets considered in this paper.
}
	\label{tab:graph_stats}
	\centering
	\scalebox{0.85}{ \renewcommand{\arraystretch}{1.0}
		
		\begin{tabular}{@{}l||A|A|A|A|A|B @{}}
			\toprule
			 & \multicolumn{6}{c}{\textbf{Dataset}} \\ 
		\cmidrule{2-7}
			& {\textsc{DD}} & {\textsc{MUTAG}} & {\textsc{MUTAG2}}
			& {\textsc{PTC-FM}} & {\textsc{PROTEINS}} & {\textsc{IMDB-B}}
			\\ \cmidrule{1-7}
			\# Graphs  & 1,168  &  188 & 4,337  &  349
			  &  1,113 & 1,000  \\ 
			\cmidrule{1-7}
			Avg. \# Nodes  &  268.71 & 17.93  & 29.76  &  14.11
			   &   39.05 &  19.77\\  
			\cmidrule{1-7}
				Avg. \# Edges  & 676.21  & 19.79  &  30.76 &  14.48
			&  72.81  & 96.53   \\  
			\cmidrule{1-7}
			\# Classes  &  2 & 2  & 2 & 2
			  &  2 & 2 \\ 
			\bottomrule
	\end{tabular}
}
\end{table*}
\begin{table*}[t]
	\caption{Some statistics of the New York City Taxi datasets.
}
	\label{tab:graph_stats2}
	\centering
	\scalebox{0.85}{ \renewcommand{\arraystretch}{1.0}
		
		\begin{tabular}{@{}l||A|A|A|A|A|B @{}}
			\toprule
			 & \multicolumn{6}{c}{\textbf{Dataset}} \\ 
		\cmidrule{2-7}
			
			& {\textsc{Jan. G.}} & {\textsc{Feb. G.}} & {\textsc{Mar. G.}}
			& {\textsc{Jan. Y.}} & {\textsc{Feb. Y.}} & {\textsc{Mar. Y.}} 
			\\ \cmidrule{1-7}
			\# Graphs 
			  & 744 & 648  & 744  & 744  & 648  & 744 \\ 
			\cmidrule{1-7}
			Avg. \# Nodes 
			   &  174.25 &  175.28 &  174.43 & 203.04  & 
			   199.28 &  207.24\\  
			\cmidrule{1-7}
				Avg. \# Edges  
		  &  497.35 & 502.96  & 480.33  & 1865.66  & 
			1868.28 & 1967.59 \\  
			\cmidrule{1-7}
			\# Classes 
			   & 2  &  2  &  2 &  2 &  2 & 2\\ 
			\bottomrule
	\end{tabular}
}
\end{table*}

\paragraph{Benchmark datasets}
These are the commonly tested binary-class datasets~\cite{morris2020tudataset} for the 
graph classification task. We use protein datasets (\textsc{DD}, \textsc{MUTAG}, 
\textsc{MUTAG2}, \textsc{PTC-FM}, and \textsc{PROTEINS}) and a collaboration network dataset \textsc{IMDB-B}. In each protein dataset, a graph is a protein, and the task is to predict whether the protein is an enzyme. In \textsc{IMDB-B}, each graph is an ego-network of actor collaborations, and the task is to predict the genre of the collaboration is either \textit{Action} or \textit{Romance}.

\paragraph{New York City Taxi datasets}
We extracted the taxi ridership data in 2019 from New York City (NYC) Taxi 
Commission\footnote{\url{https://www1.nyc.gov/site/tlc/about/tlc-trip-record-data.page}}. The areas in New York are represented as 
nodes, and each taxi trip is an edge connecting the source and destination 
nodes. All taxi trips in an 1-hour interval form a graph, and each dataset 
spans 
a month of taxi operations. We augmented the binary label for each graph 
as taxi trips in weekdays (Mon-Thu) vs. weekend (Fri-Sun). We considered two 
taxi 
operators (Yellow and Green) and processed data in January, February and March 
of 2019, making 6 datasets in total: \textsc{Jan. G.}, \textsc{Feb. G.}, \textsc{Mar. G.}, \textsc{Jan. Y.}, \textsc{Feb. Y.}, and \textsc{Mar. Y.}.
 
\subsubsection{Experimental procedure}
\label{subsubsec:procedure}
We tested the ability of each GNN architecture to classify graphs in the 
following settings:
\begin{itemize}
	\item Original setting: The GNN with a final classifier outputs the 
	estimated class 
	probabilities, and the weights are updated by the cross-entropy loss 
	with respect to the ground truth. We use the implementation 
	provided by the authors. To enhance the capacity of the final classifier, 
	we tune it by using up to three fully-connected layers and 
	select the model based on validation sets.
	\item \ourtriplet and \ourfinetune: See Section~\ref{sec:method}.
\end{itemize}

In addition, we compared our two-stage method with the transfer-learning 
methods in~\cite{hlugzlplstrategies} and~\cite{lu2021learning}, which also claimed the effectiveness of a 
pre-training strategy. Out of the 5 GNN models investigated in our work, 
\textsc{GraphSage} and \textsc{GAT} were provided with trained weights by Hu et al.~\cite{hlugzlplstrategies}, and they were compared with \textsc{GraphSage} and \textsc{GAT} trained in 
\ourfinetune. We also pre-trained \textsc{GraphSage} and \textsc{GAT} following the procedure in~\cite{lu2021learning} 
and then fine-tuned them on our graph classification task and compared against \ourfinetune.

Each dataset was randomly split into three sets: training (80\%), validation 
(10\%) and test (10\%) sets.
The reported results are the mean and standard deviation of test accuracy over five splits.

We initialized node features as learnable features that were 
optimized with GNN parameters during training. 
While input features are provided in some datasets, according to our preliminary study, using learnable features often led to better accuracy.

In \ourtriplet and \ourfinetune, each graph was used as an anchor once in each 
iteration, while the positive and negative graphs were chosen randomly. 
While Schroff et al.~\cite{schroff2015facenet} suggested the potential advantage of choosing 
``hard'' triplets and ``semi-hard'' triplets, what we empirically found was that the two 
options did not improve the classification accuracy. We used multi-layer 
perceptron (MLP) as the classifier.

\subsubsection{Hyperparameter search}
\label{subsubsec:hyperparameter}
For each GNN, the hyperparameters regarding the network 
architecture were tuned in the 
same search space for each of the three settings to ensure fairness: original, 
\ourtriplet, and
\ourfinetune. The search space for the dimensions of the input vector, hidden 
vector and output vector for all GNNs was $\{16, 32, 64, 96, 128\}$. For 
\textsc{DiffPool}, we used three 
layers of graph convolution and one \textsc{DiffPool} layer as described in the original 
paper. For \textsc{EigenGCN}, we used three pooling operators as it was shown to 
achieve 
the best performance in the original paper. For \textsc{SAGPool}, we used  three 
pooling 
layers as explained in the original paper. Other hyperparameters that are 
exclusive to each GNN architecture were set to the
default values provided in each paper's original code of each architecture's 
authors. 

In each of the three settings, the architecture of the final classifier was 
also tuned in the same search space. 
The number of fully-connected layers was up to 3 while the search space for 
the hidden dimension was $\{2^h | 1 \leq h \leq \log_2(d)\}$, where $d$ is the dimension of the output vector. As shown in Appendix~\ref{appendix:hyperparameter}, using a strong classifier is helpful in improving the classification accuracy.

The two settings \ourtriplet and \ourfinetune require an additional 
hyperparameter $\alpha$ as the margin in the triplet loss. 
While Schroff et al.~\cite{schroff2015facenet} found $\alpha=0.2$ to 
be effective, we empirically found that this value was too small to separate 
instances of different classes. Instead, the search space for $\alpha$ we used 
was $\{0.5, 1.0, 1.5, 2.0, 2.5\}$. Grid search showed that the performance was 
sensitive with respect to the choice of $\alpha$.

\subsubsection{Computing specifications \& dependencies}
All experiments were performed on Ubuntu 18.04 LTS running on a machine with 4 
RTX-2080Ti GPUs, each of which has $11$GB memory. The following libraries were 
used to run the code:
 torch 1.7.0, networkx 2.5,  sklearn 0.23.2, numpy 1.16.4, and torch-geometric 1.6.3.

\subsection{Results and discussion}

\subsubsection{Improvement by our methods}
We draw some observations from the results of comparing \ourtriplet and \ourfinetune with the original 
setting (Tables~\ref{tab:triplet_ppi} 
and~\ref{tab:triplet_taxi}):

\begin{table*}[htbp!]	
	\caption{Average and standard deviation (in \%) of graph classification accuracies in benchmark datasets in three 
	settings: 
	original, \ourtriplet, and \ourfinetune. 
	Pre-training 
	GNNs in \ourtriplet 
	improves the 
	classification accuracy compared to the original setting, and fine-tuning GNNs in \ourfinetune further improves the accuracy.
	 The average gain is in \% points.}
	\label{tab:triplet_ppi}
	\centering
	\scalebox{0.85}{\renewcommand{\arraystretch}{1.0}
		
		\begin{tabular}{@{}l||c|c|c|c|c|c||c@{}}
			\toprule
			\textbf{Method} & \multicolumn{6}{c}{\textbf{Dataset}} & \textbf{Average Gain} \\ 
			\cmidrule{2-7}
			& {\textsc{DD}} & {\textsc{MUTAG}} & {\textsc{MUTAG2}} & 
			{\textsc{PTC-FM}} &  {\textsc{PROTEINS}} & {\textsc{IMDB-B}} & (in \% points)
			\\ \cmidrule{1-8}
			\textsc{GraphSage}  & 69.24 $\pm$ 0.52 & 65.13 $\pm$ 0.87    & 75.44 $\pm$ 0.50 & 
			61.77 $\pm$ 1.11 & 71.25 $\pm$ 1.38 & 65.52 $\pm$ 0.96 & - \\ 
			\textsc{GraphSage (\ourtriplet)} & 75.13 $\pm$ 0.82  & 
			80.86 $\pm$ 
			1.19 & 76.84 $\pm$ 0.54   & \textbf{62.75 $\pm$ 1.20}  & 71.29 $\pm$ 
			0.41 & 
			\textbf{68.37 $\pm$ 0.63} & 4.47
			\\     
			\textsc{GraphSage (\ourfinetune)} &  \textbf{76.52 $\pm$ 1.47} &  
			\textbf{81.14 $\pm$	0.68} &  \textbf{77.71 $\pm$ 0.41}  & 62.65 $\pm$ 0.72 & \textbf{72.34 $\pm$ 
			0.56} & 
			68.24 $\pm$ 0.83  &  5.01\\     
			\cmidrule{1-8}
			\textsc{GAT}  & 66.50 $\pm$ 1.24  & 65.18 $\pm$ 1.03  &  76.23 $\pm$ 0.67   & 60.65 $\pm$ 0.42 
			& 66.92 $\pm$ 0.75  & 67.13 $\pm$ 0.88 & -\\ 
			\textsc{GAT (\ourtriplet)} & 72.95 $\pm$ 0.91 &  77.84 
			$\pm$ 0.63 &  
			76.34 $\pm$ 0.52 & \textbf{62.04 $\pm$ 1.16} & 70.17 $\pm$ 0.72 & 
			\textbf{69.15 $\pm$ 0.87} & 4.11\\     
			\textsc{GAT (\ourfinetune)} & \textbf{74.13 $\pm$ 1.47} & 
			\textbf{78.17 
			$\pm$ 1.41} & 
			\textbf{76.49 $\pm$ 1.23}   & 61.61 $\pm$ 0.53 & \textbf{72.64 
			$\pm$ 0.58} & 
			67.25 $\pm$ 0.89 & 5.37\\     
			\cmidrule{1-8}
			\textsc{DiffPool}  & 72.11 $\pm$ 0.42 & 86.32 $\pm$ 0.83 & 77.21 $\pm$ 1.16  & 61.15 $\pm$ 0.35   & 72.24 $\pm$ 0.67 & 64.93 $\pm$ 0.74 & -\\ 
			\textsc{DiffPool (\ourtriplet)}  & 74.93 $\pm$ 0.53 & 
			86.14 $\pm$ 
			0.77 &  \textbf{77.94 $\pm$ 1.28} & 62.03 $\pm$ 0.32 & \textbf{73.87 $\pm$ 
			0.64} & 
			\textbf{65.22 $\pm$ 0.83} & 1.03 \\     
			\textsc{DiffPool (\ourfinetune)} & \textbf{78.84 $\pm$ 0.54} & 
			\textbf{87.38 $\pm$	0.62} & 77.08 $\pm$ 1.23 & \textbf{62.15 $\pm$ 0.68}   & 73.07 $\pm$ 
			1.17 
			&64.90 $\pm$ 0.81 & 1.07 \\     
			\cmidrule{1-8}
			\textsc{EigenGCN}  & 75.62 $\pm$ 0.63 & 79.87 $\pm$ 0.66 & 76.65 $\pm$ 1.14  & 63.34 $\pm$ 1.23 & 75.63 $\pm$ 0.82 & 71.86 $\pm$ 0.55 & -\\ 
			\textsc{EigenGCN (\ourtriplet)}  & 77.56 $\pm$ 0.48 & 
			80.21 $\pm$ 
			0.71 & \textbf{77.98 $\pm$ 0.62} &  \textbf{64.13 $\pm$ 0.95} &  75.93 $\pm$ 
			0.56  & 
			\textbf{72.66 $\pm$ 0.42} & 0.91 \\     
			\textsc{EigenGCN (\ourfinetune)} & \textbf{78.13 $\pm$ 0.51} & 
			\textbf{81.42 $\pm$	0.86} & 77.02 $\pm$ 1.72  & 63.52 $\pm$ 1.43  & \textbf{77.31 $\pm$ 1.46} & 72.04 
			$\pm$ 0.53 & 1.07
			\\     
			\cmidrule{1-8}
			\textsc{SAGPool}  & 76.12 $\pm$ 0.79 & 78.34 $\pm$ 0.65 & 76.83 $\pm$ 1.27  & 63.27 $\pm$ 0.78 & 74.34 $\pm$ 1.25 & 71.23 $\pm$ 1.12 & -\\ 
			\textsc{SAGPool (\ourtriplet)}  & \textbf{78.32 $\pm$ 1.26} & 
			\textbf{79.63 $\pm$	0.95} & \textbf{78.03 $\pm$ 0.68} &  63.83 $\pm$ 0.83 & \textbf{77.52 $\pm$	0.54} & 71.73 
			$\pm$ 0.81 & 1.48
			\\     
			\textsc{SAGPool (\ourfinetune)} & 78.22 $\pm$ 0.70 & 
			79.03 $\pm$ 
			0.89 & 77.03 $\pm$ 0.63  & \textbf{64.34 $\pm$ 0.86}  & 76.23 $\pm$ 
			1.12 & \textbf{72.36 $\pm$ 0.73} & 1.24
			\\     
			\bottomrule
	\end{tabular}}
\end{table*}

\begin{itemize}
	\item Pre-training using triplet loss (i.e., the first stage of \ourtriplet and \ourfinetune)
	consistently enhances the graph classification accuracy
	of each GNN model by $0.9-5.4\%$ 
	points\footnote{For example, when the accuracy of the original setting is $63\%$ and that of our method is $67\%$, our method improves the accuracy by $4\%$ points. The accuracy improvement is $67\% - 63\% = 4\%$, so, with respect to the original setting’s accuracy (which is $63\%$), this is an improvement of: $4/63 \approx 6.3\%$. Since our focus is on the absolute accuracy of classification, we claim ``$4\%$ points” instead of ``$6.3\%$”.}, compared to its original setting.
	\item Fine-tuning the weights of GNNs (i.e, the second stage of \ourfinetune) further improves the 
	accuracy from \ourtriplet in some cases by up to 1.3\% points.
\end{itemize}

\begin{table*} [t]
	\caption{Average and standard deviation (in \%) of graph classification accuracies in NYC Taxi datasets in three 
	settings: 
		original, \ourtriplet, and \ourfinetune. Similarly 
		to the case of the benchmark datasets, \ourtriplet and \ourfinetune 
		significantly 
		outperform the original setting. \textsc{G., Y.} stand for 
		\textsc{Green, Yellow}. The average gain is in  \% points.}
	\label{tab:triplet_taxi}
	\centering
	\scalebox{0.85}{ \renewcommand{\arraystretch}{1.0}
		
		\begin{tabular}{@{}l||c|c|c|c|c|c||c @{}}
			\toprule
			\textbf{Method} & \multicolumn{6}{c}{\textbf{Dataset}} & \textbf{Average Gain}\\ 
		\cmidrule{2-7}
			
			& {\textsc{Jan. G.}} & {\textsc{Feb. G.}} & {\textsc{Mar. 
			G.}} & {\textsc{Jan. Y.}} & {\textsc{Feb. Y.}} & 
			{\textsc{Mar. Y.}} & (in \% points)
			\\ \cmidrule{1-8}
			\textsc{GraphSage}  & 73.14 $\pm$ 0.62  & 66.35 $\pm$ 1.25  & 64.63 
			$\pm$ 0.83  & 
			72.86 $\pm$ 0.92  & 64.37 $\pm$ 0.87  & 68.12 $\pm$ 0.76 & -\\ 
			\textsc{GraphSage (\ourtriplet)} & 76.14 $\pm$ 0.93 & 66.67 $\pm$ 
			1.31 & 67.13 $\pm$ 0.85 
			&  \textbf{75.24 $\pm$ 1.16}  & 65.43 $\pm$ 0.68  &  70.15 $\pm$ 0.64 & 
			1.88\\     
			\textsc{GraphSage (\ourfinetune)} & \textbf{76.63 $\pm$ 0.82}  & \textbf{67.74 $\pm$ 
			0.88}  & \textbf{68.95 $\pm$ 1.41} & 
			75.21 $\pm$ 1.70  & \textbf{67.64 $\pm$ 0.73}  &  \textbf{70.23 $\pm$ 1.25} & 2.82 
			\\     
			\cmidrule{1-8}
			\textsc{GAT}  & 71.26 $\pm$ 1.51  & 67.82 $\pm$ 0.77  & 66.13 $\pm$ 
			0.72  & 72.64 $\pm$ 0.54 
			 & 64.76 $\pm$ 0.73  & 67.51 $\pm$ 1.69 & - \\ 
			\textsc{GAT (\ourtriplet)} & 75.23 $\pm$ 0.82 & 67.24 $\pm$ 0.56 & 
			67.34 $\pm$ 0.71 &  \textbf{76.82 $\pm$ 1.23} & 66.45 $\pm$ 0.85 &  70.66 
			$\pm$ 
			0.78 & 2.27
			\\     
			\textsc{GAT (\ourfinetune)} & \textbf{74.65 $\pm$ 0.98} & 
			\textbf{68.11 $\pm$ 0.69} & 
			\textbf{69.15 $\pm$ 1.37}  & 74.79 $\pm$ 1.27  & \textbf{68.75 
			$\pm$ 0.66} & \textbf{70.44 
			$\pm$ 
			0.93}  & 3.04
			\\     
			\cmidrule{1-8}
			\textsc{DiffPool}  & 78.43 $\pm$ 0.74 & 73.12 $\pm$ 0.42 & 71.39 
			$\pm$ 1.56  & 72.52 $\pm$ 1.23 & 67.43 $\pm$ 0.87 & 74.34 $\pm$ 
			0.77 & -\\ 
			\textsc{DiffPool (\ourtriplet)}  & \textbf{80.28 $\pm$ 1.16} & \textbf{75.69 $\pm$ 
			1.21} & \textbf{73.79 $\pm$ 0.81} & 75.09 $\pm$ 0.72  & 68.19 $\pm$ 0.50 &  
			74.87 $\pm$ 0.83 & 1.78\\     
			\textsc{DiffPool (\ourfinetune)} & 79.63 $\pm$ 0.82 & 74.56 $\pm$ 
			1.32 & 72.92 $\pm$ 0.65  & \textbf{75.95 $\pm$ 1.21}  & \textbf{69.31 $\pm$ 0.97} & 
			\textbf{75.76 
			$\pm$ 0.86} & 1.81
			\\     
			\cmidrule{1-8}
			\textsc{EigenGCN}  & 75.45 $\pm$ 0.44 & 69.32 $\pm$ 1.82 & 72.21 
			$\pm$ 0.83  & 73.21 $\pm$ 1.35 & 69.64 $\pm$ 0.76 & 69.52 $\pm$ 1.54 
			& -\\ 
			\textsc{EigenGCN (\ourtriplet)}  & \textbf{77.14 $\pm$ 0.81} & 70.03 $\pm$ 
			0.62 & \textbf{74.12 $\pm$ 1.34} &  74.36 $\pm$ 1.65 & 69.72 $\pm$ 0.97 &  
			70.03 $\pm$ 0.86  & 1.02
			\\     
			\textsc{EigenGCN (\ourfinetune)} & 76.73 $\pm$ 1.21 & \textbf{71.27 $\pm$ 
			1.33} & 73.37 $\pm$ 1.85  & \textbf{75.33 $\pm$ 1.14}  & \textbf{71.65 $\pm$ 1.67} & 
			\textbf{71.84 
			$\pm$ 0.62} & 1.80
			\\     
			\cmidrule{1-8}
			\textsc{SAGPool}  & 73.23 $\pm$ 0.59 & 67.46 $\pm$ 0.73  & 72.78 
			$\pm$ 1.34   & 
			72.65 $\pm$ 0.72 & 68.83 $\pm$ 1.25 & 69.68 $\pm$ 1.35 & - \\ 
			\textsc{SAGPool (\ourtriplet)}  & \textbf{76.36 $\pm$ 1.37} & 69.07 $\pm$ 
			1.48 & \textbf{74.34 $\pm$ 1.52} & 71.11 $\pm$ 0.73  & \textbf{70.02 $\pm$ 0.64} &  
			70.04 $\pm$ 1.48 & 1.05
			\\     
			\textsc{SAGPool (\ourfinetune)} & 75.38 $\pm$ 0.86 & \textbf{69.27 $\pm$ 
			1.12} & 73.19 $\pm$ 1.34   & \textbf{72.51 $\pm$ 0.85}  & 69.16 $\pm$ 0.79 & 
			\textbf{70.59 $\pm$ 0.52} & 0.91
			\\    
			\bottomrule
	\end{tabular}
}
\end{table*}

\begin{figure}[htbp]
	\addtolength{\tabcolsep}{-3pt}
	\begin{center}
		\begin{subfigure}{0.32\textwidth}
		    \centering
			\includegraphics[width=1.85in]{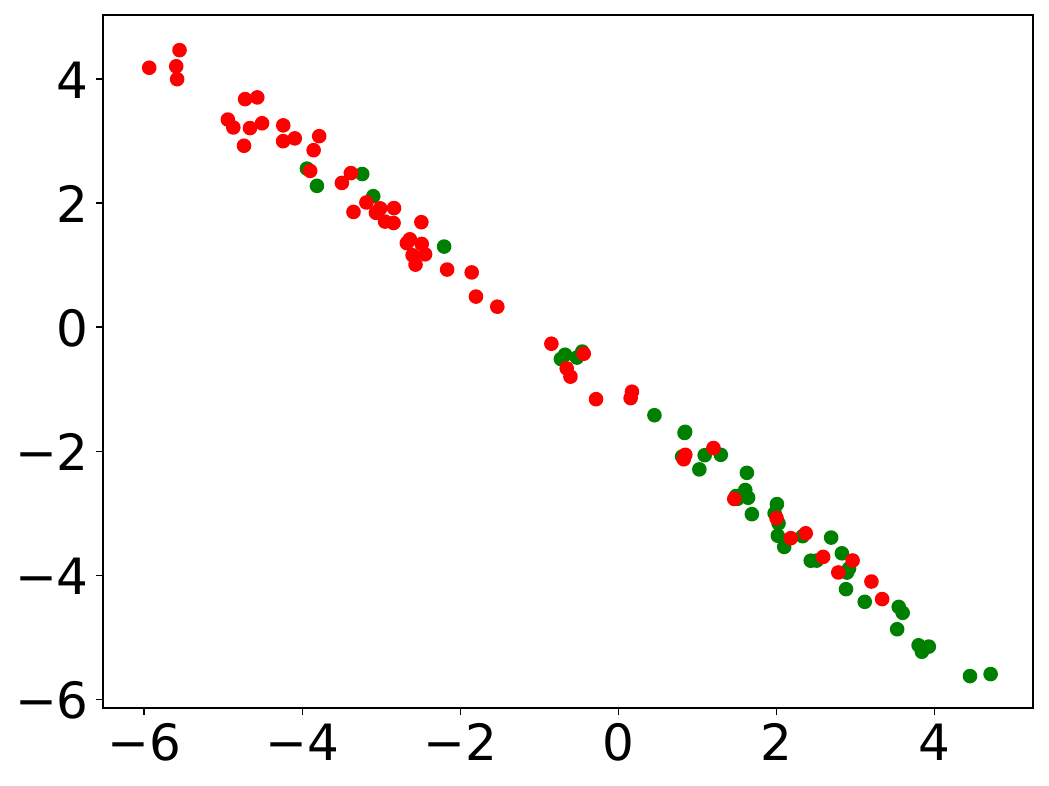} 
			\caption{\textsc{Original}}
		\end{subfigure}
		\begin{subfigure}{0.32\textwidth}
		\centering
			\includegraphics[width=1.93in]{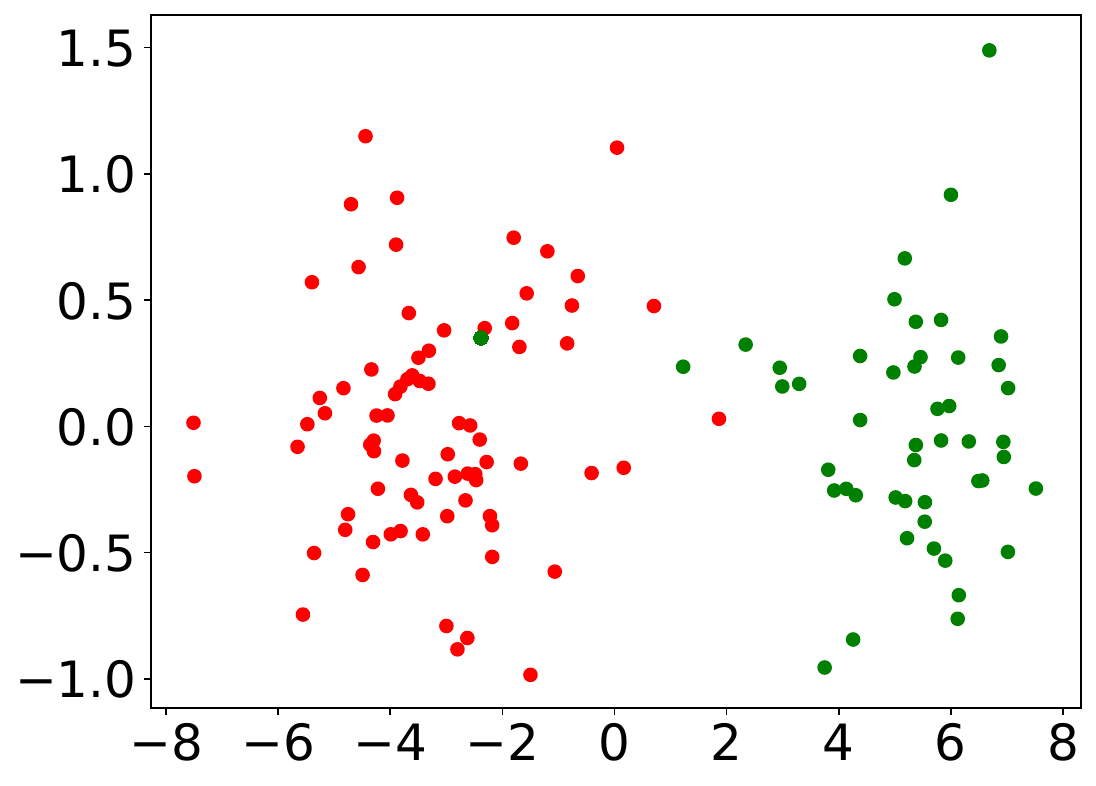}
			\caption{\textsc{\ourtriplet}}
		\end{subfigure}
		\begin{subfigure}{0.32\textwidth}
		\centering
			\includegraphics[width=1.93in]{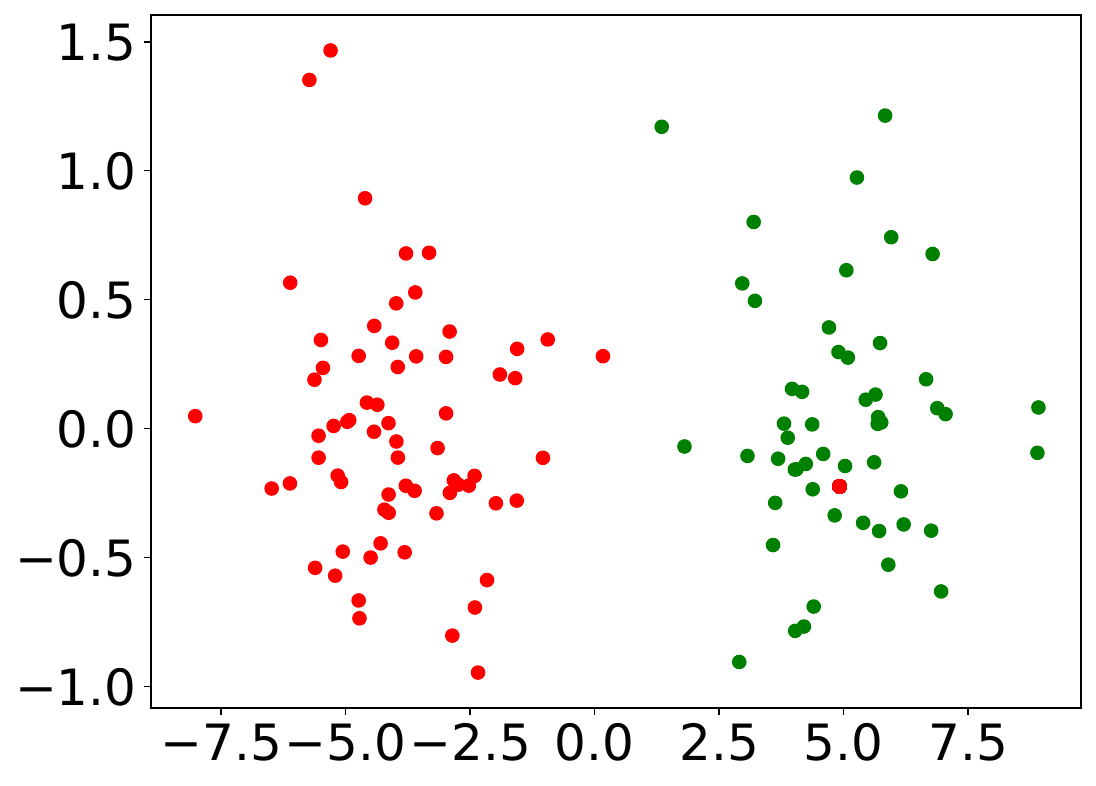}
			\caption{\textsc{\ourfinetune}}
		\end{subfigure}
		\caption{\label{fig:dd}
			Visualization of the final embeddings in Original (left), 
			\ourtriplet (middle) and \ourfinetune (right) settings (\textsc{DD} 
			dataset). 
			Instances of the two classes are separated better in \ourtriplet 
			and \ourfinetune
			than in the 
			original setting.
		}	
	\end{center}
\end{figure}

\begin{figure}[htbp!]
	\addtolength{\tabcolsep}{-3pt}
	\begin{center}
	    \textbf{\textsc{DD} Dataset:} \hfill \ \  \\
		\begin{subfigure}{0.19\textwidth}
		    \centering
			\includegraphics[width=1.3in]{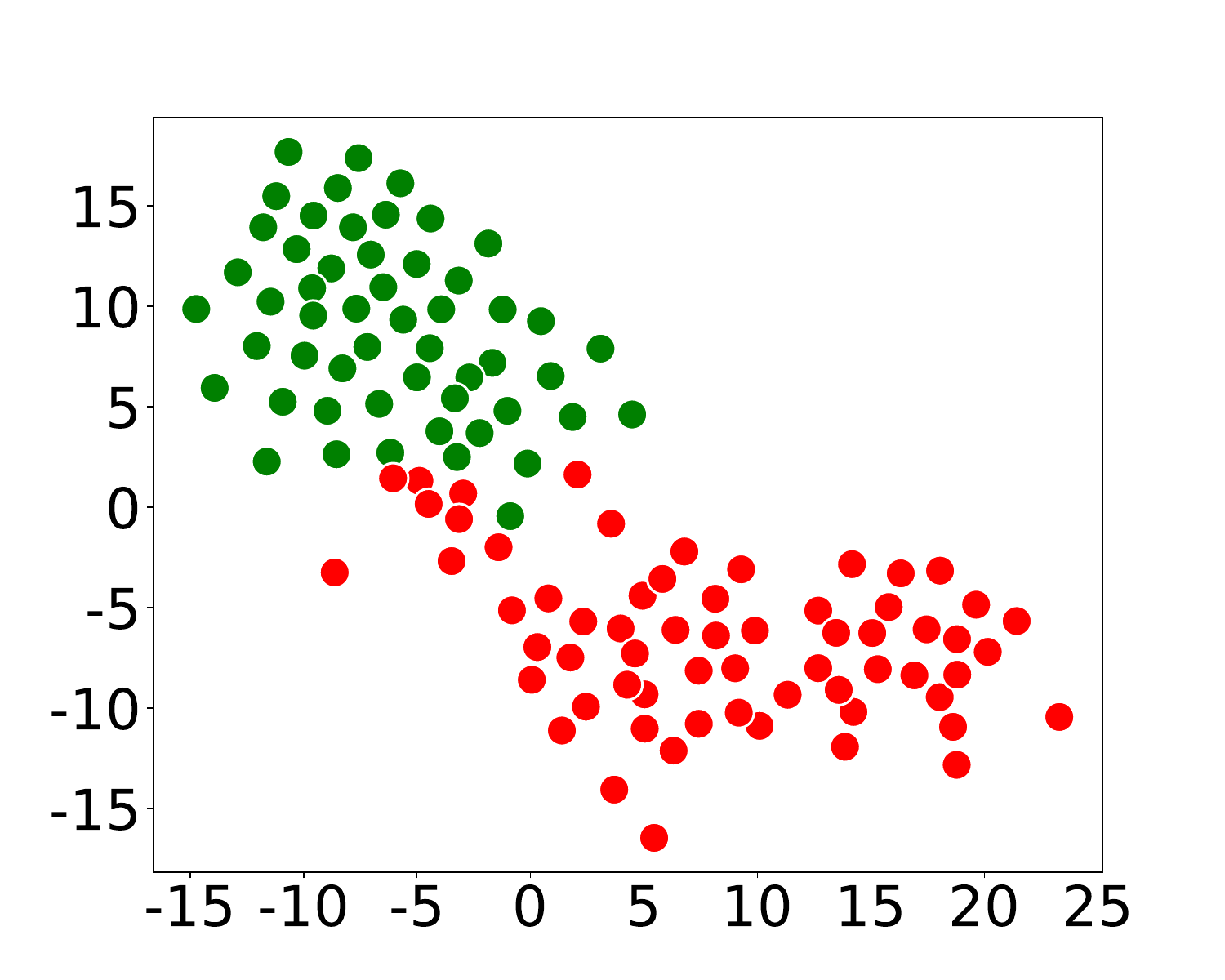} 
		\end{subfigure}
		\begin{subfigure}{0.19\textwidth}
		    \centering
			\includegraphics[width=1.3in]{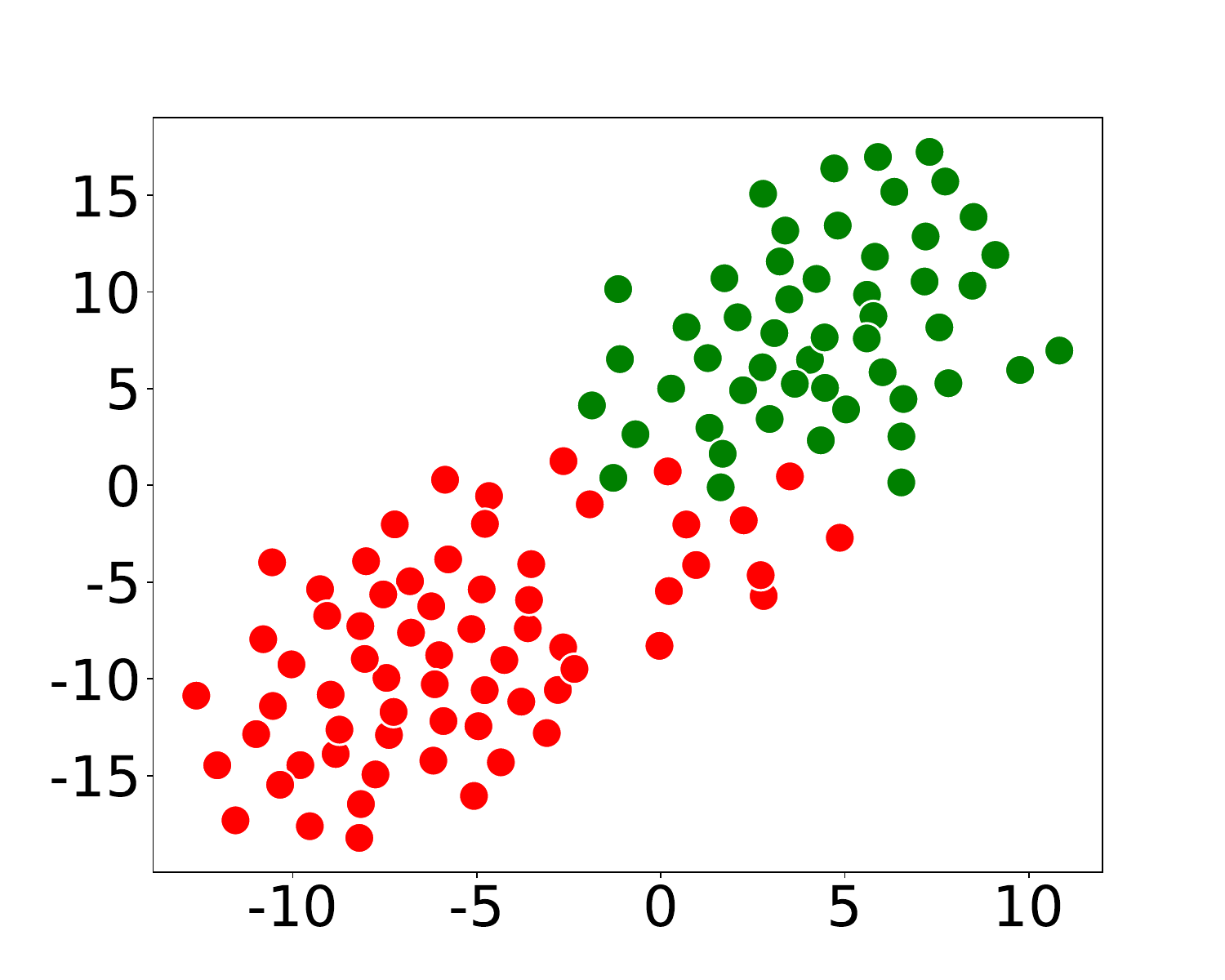}
		\end{subfigure}
		\begin{subfigure}{0.19\textwidth}
		    \centering
			\includegraphics[width=1.3in]{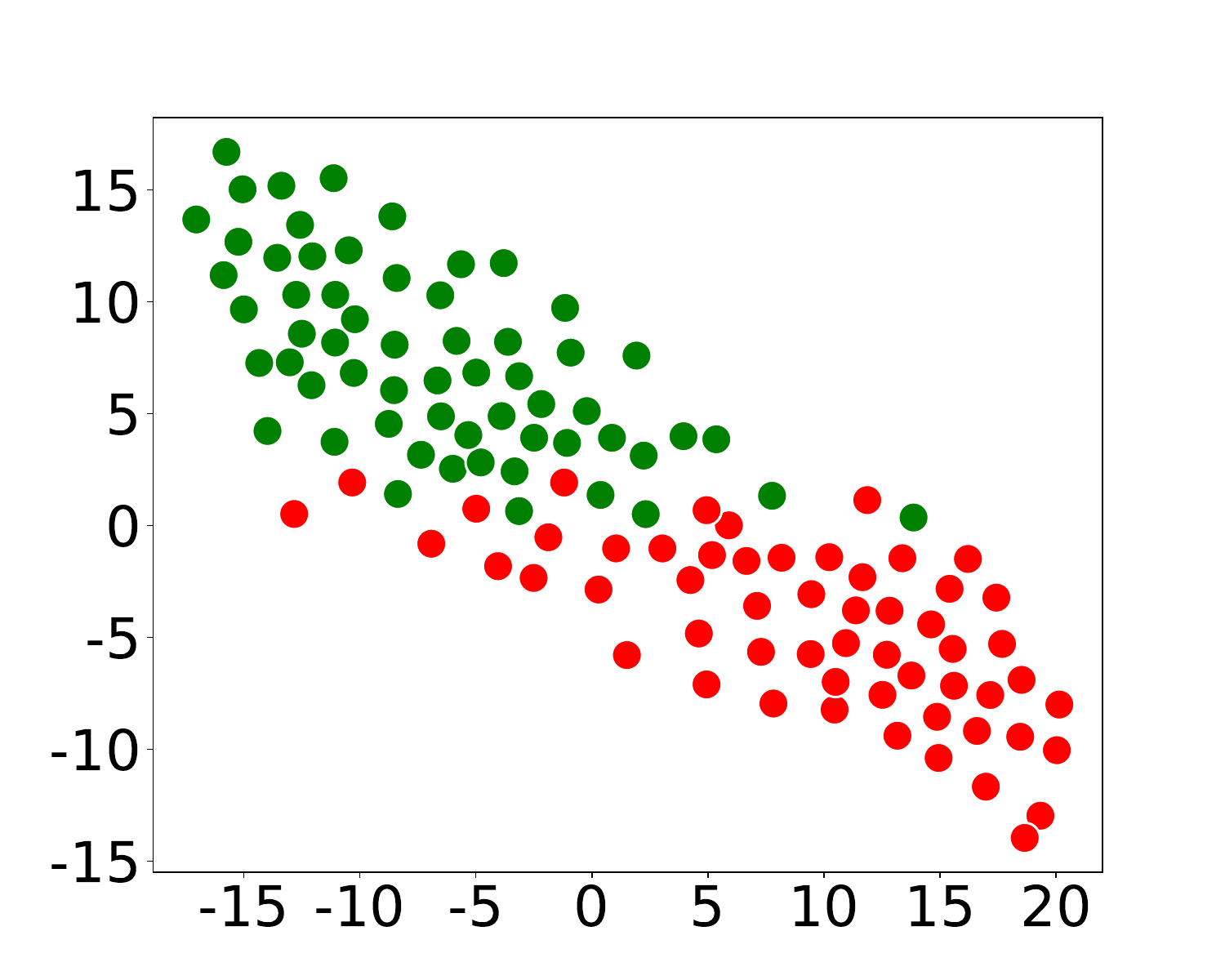}
		\end{subfigure}
		\begin{subfigure}{0.19\textwidth}
		    \centering
			\includegraphics[width=1.3in]{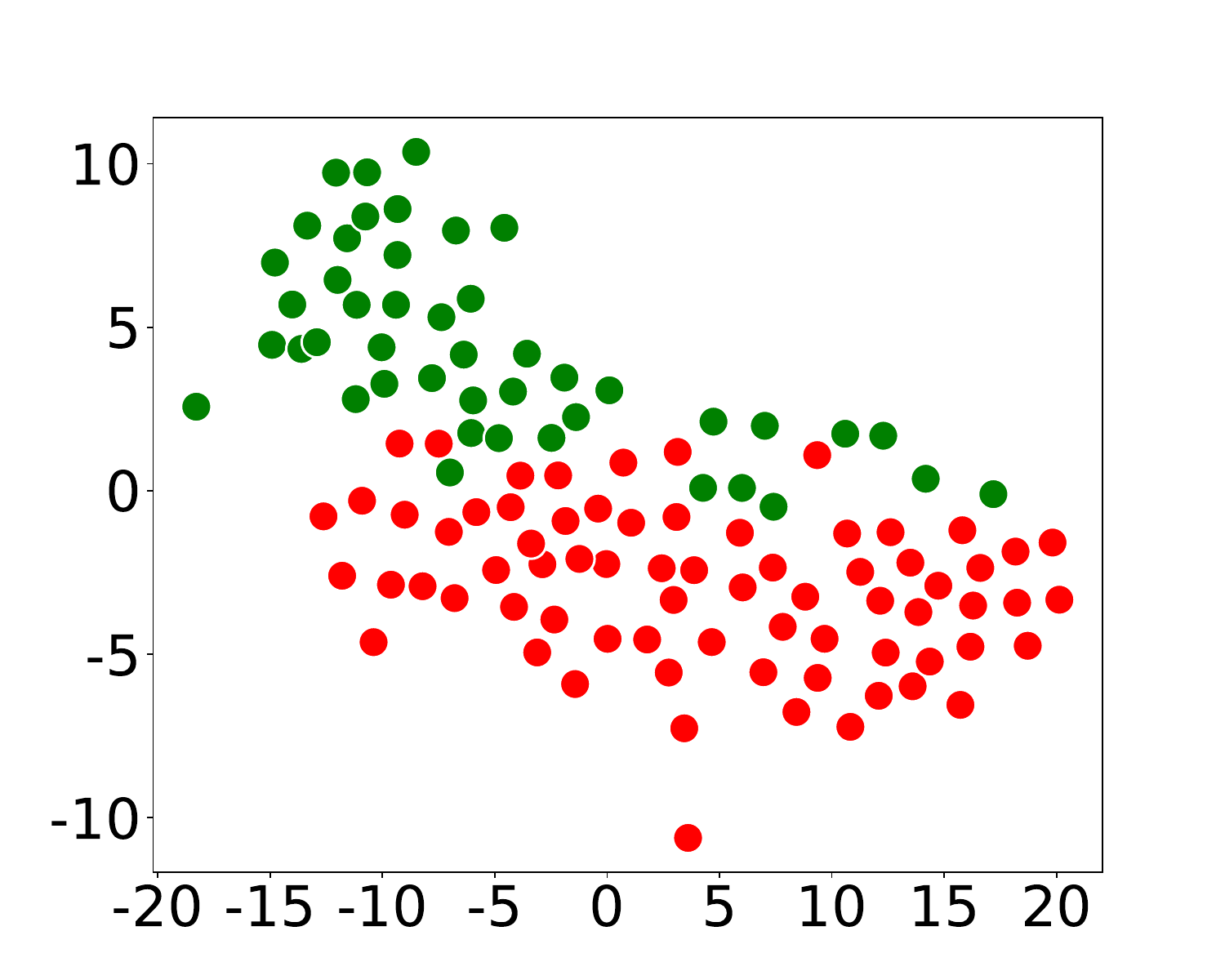}
		\end{subfigure}
		\begin{subfigure}{0.19\textwidth}
		    \centering
			\includegraphics[width=1.3in]{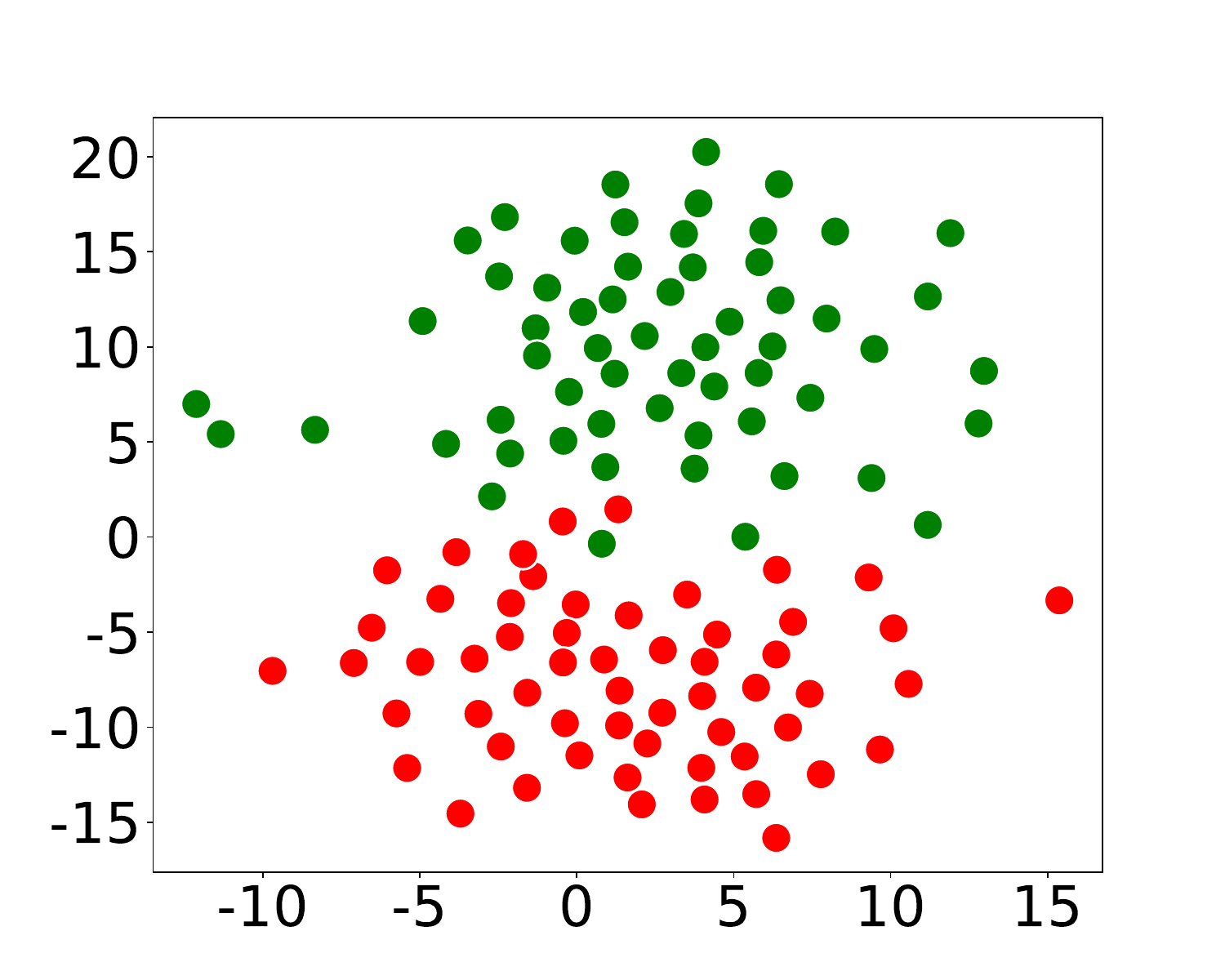}
		\end{subfigure} \\
		\vspace{1mm}
		\textbf{\textsc{Feb. G.} dataset:} \hfill \ \  \\
        \begin{subfigure}{0.19\textwidth}
            \centering
			\includegraphics[width=1.3in]{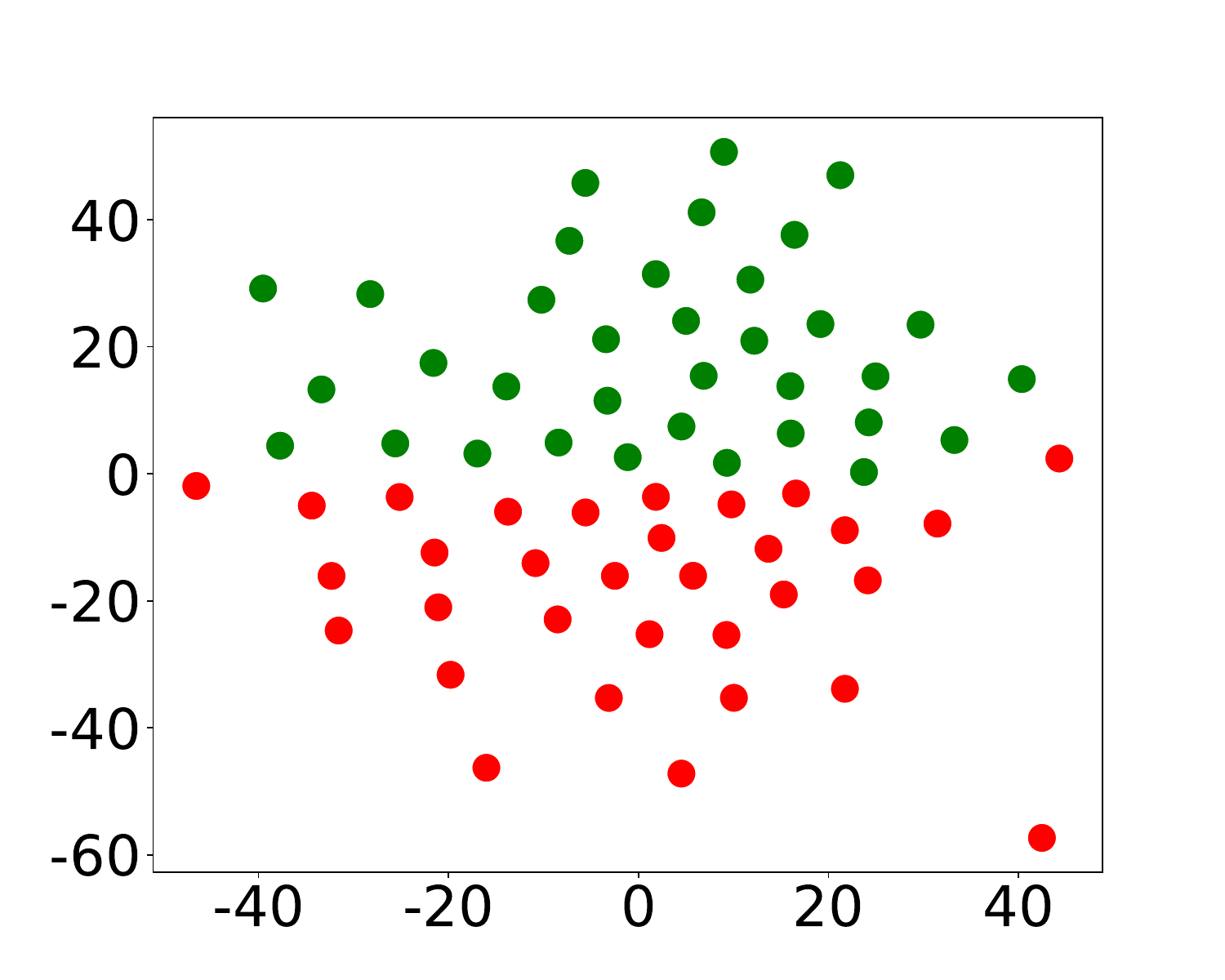} 
			\caption{\textsc{GraphSage}}
		\end{subfigure}
		\begin{subfigure}{0.19\textwidth}
		    \centering
			\includegraphics[width=1.3in]{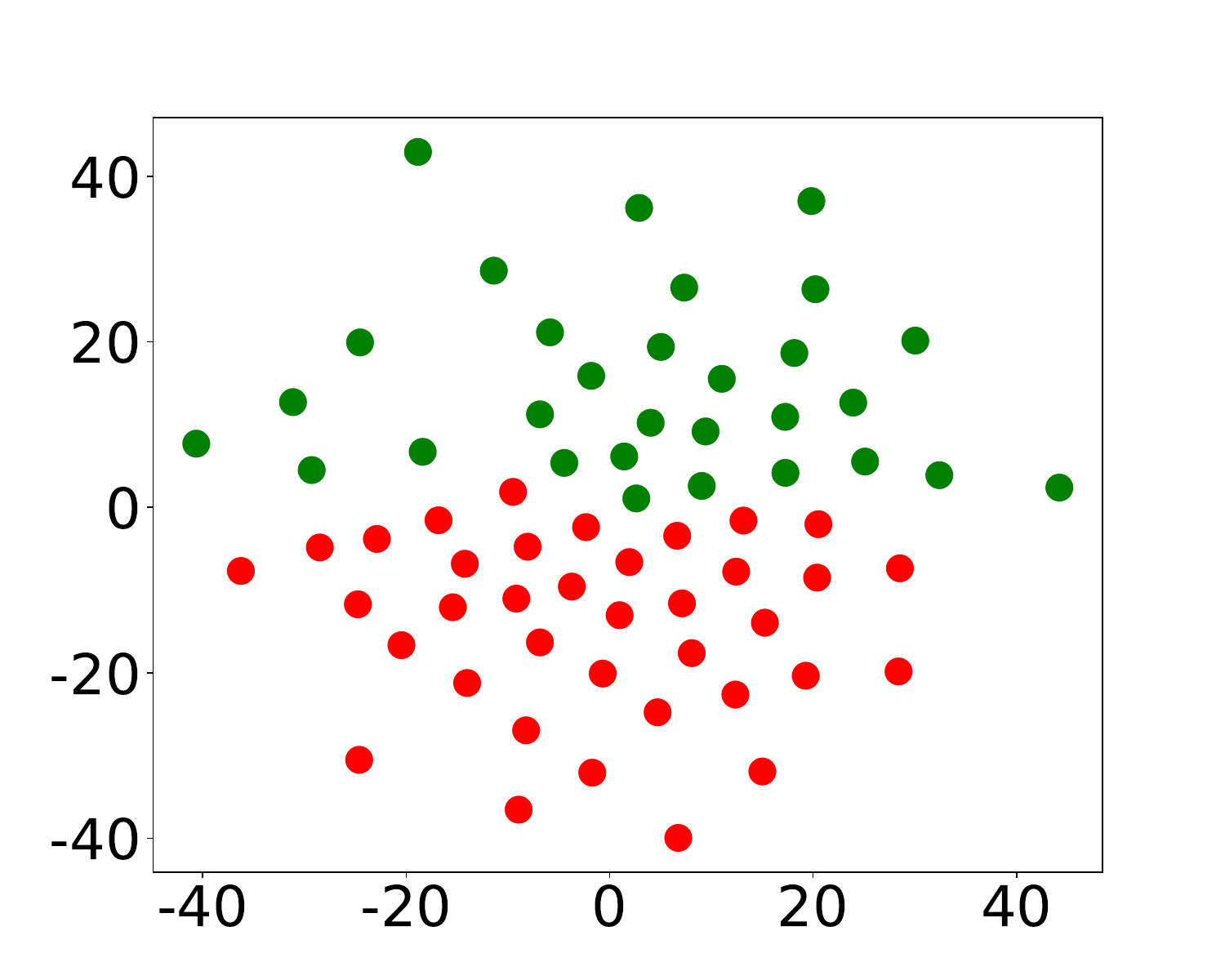}
			\caption{\textsc{GAT}}
		\end{subfigure}
		\begin{subfigure}{0.19\textwidth}
		    \centering
			\includegraphics[width=1.3in]{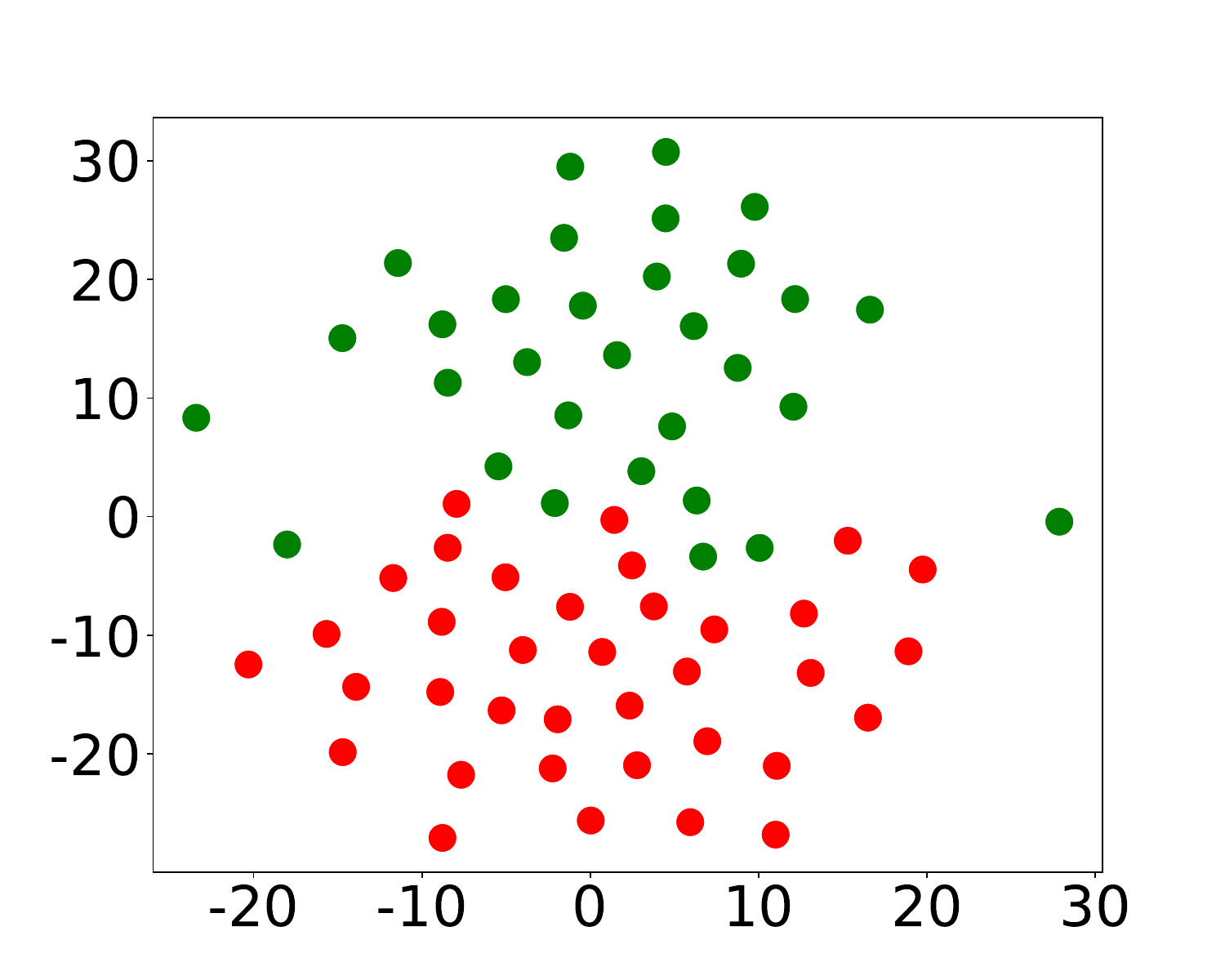}
			\caption{\textsc{DiffPool}}
		\end{subfigure}
		\begin{subfigure}{0.19\textwidth}
		    \centering
			\includegraphics[width=1.3in]{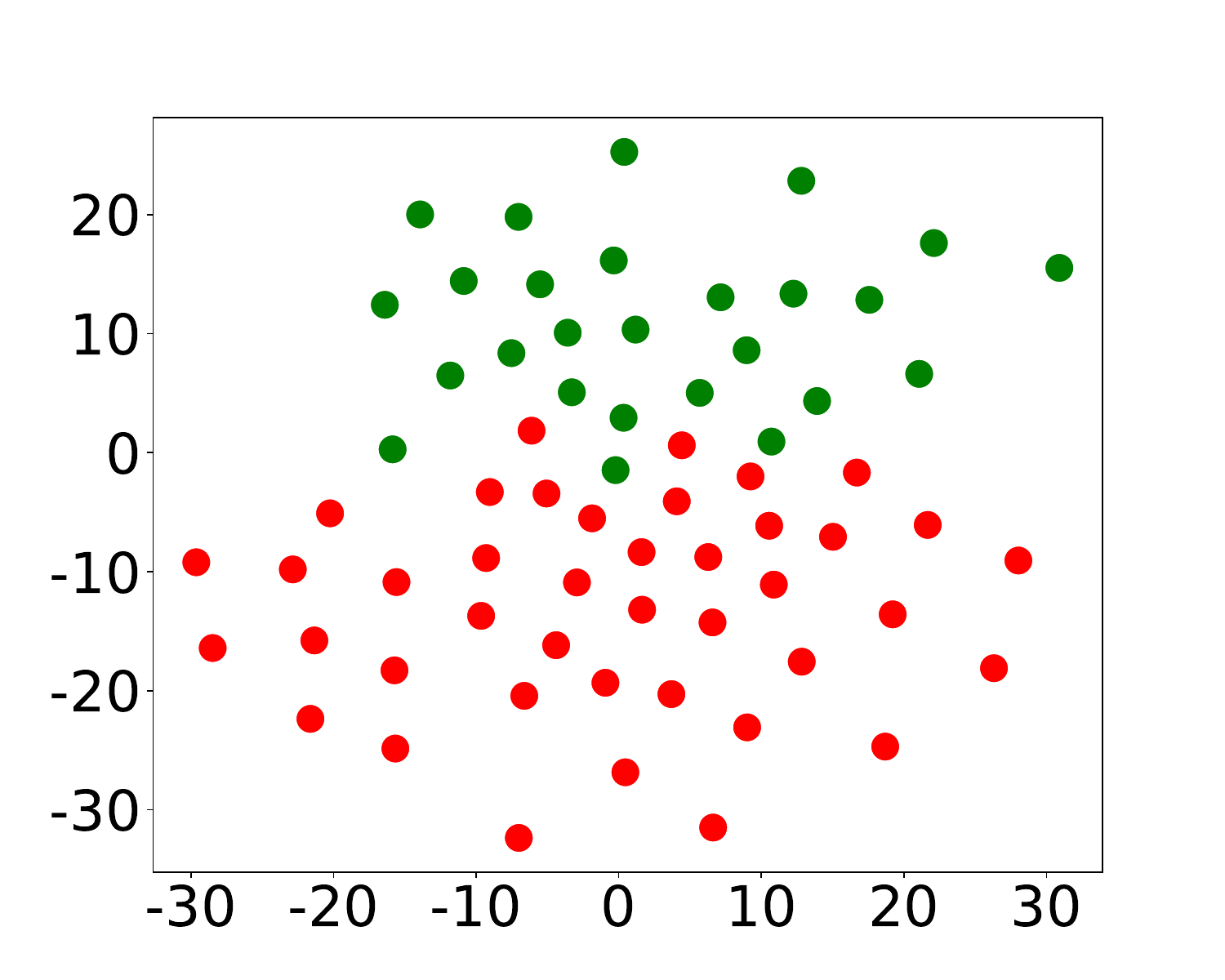}
			\caption{\textsc{EigenGCN}}
		\end{subfigure}
		\begin{subfigure}{0.19\textwidth}
		    \centering
			\includegraphics[width=1.3in]{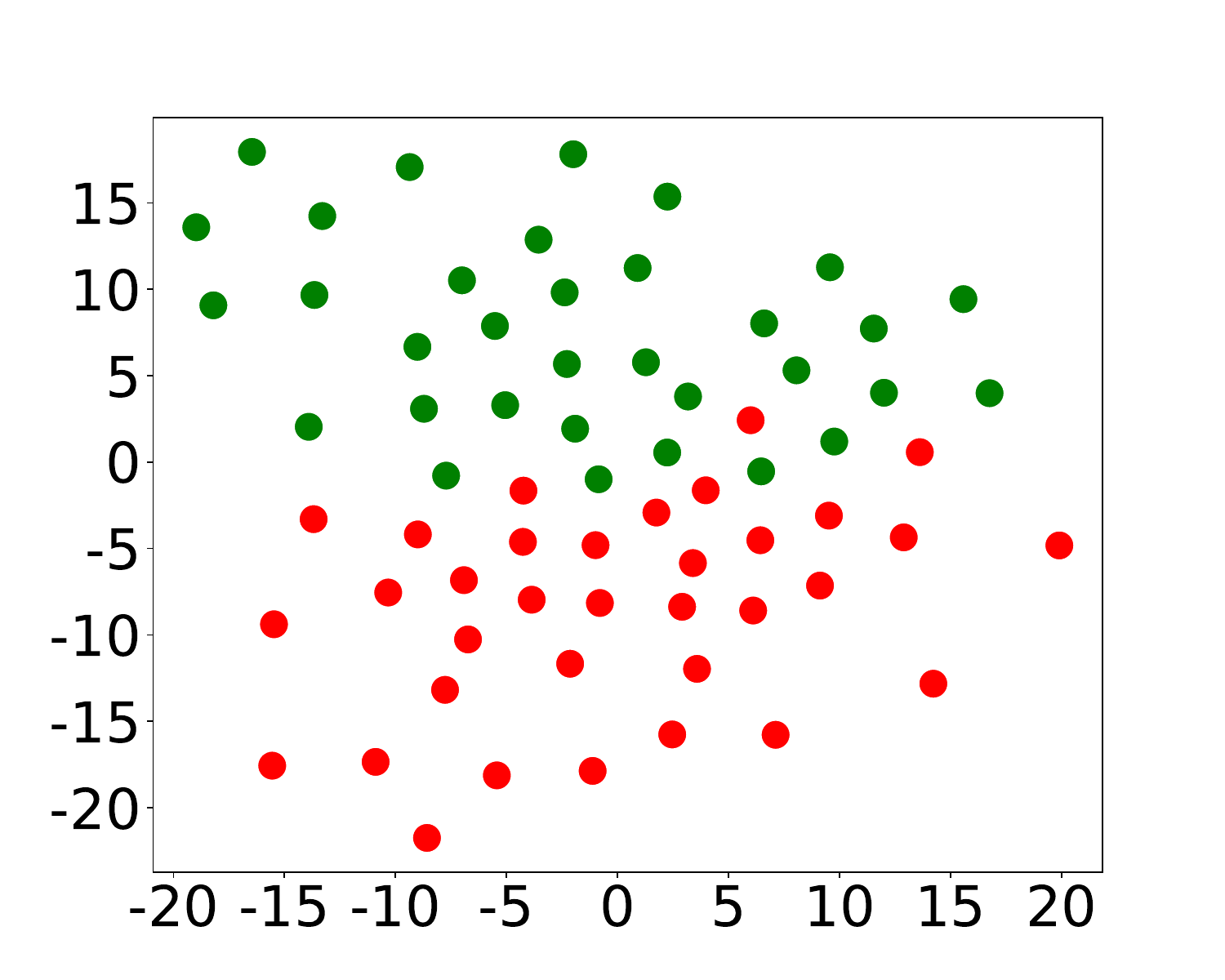}
			\caption{\textsc{SAGPool}}
		\end{subfigure}		
		\caption{\label{fig:tsne}
			Visualization of the graph embeddings after training with the triplet loss in the \textsc{DD} (top) and \textsc{Feb. G.} (bottom)
			datasets. T-SNE is applied to reduce the embedding dimension from 64 to 2 for visualization. Two colors represent two classes. Training with the triplet loss separates most instances of the two classes.
		}	
	\end{center}
\end{figure}

The improvement in accuracy by \ourtriplet and \ourfinetune suggests two 
explanations:
\begin{itemize}
	\item The end-to-end training methods fail to realize the full potential 
	of the GNN models. Even if the final classifier is 
	upgraded from a fully-connected layer to an MLP, the accuracy is not as high 
	as in \ourtriplet and \ourfinetune.
	\item Learning meaningful embeddings in between that are fairly separated 
	based on classes (see Figures~\ref{fig:dd} and~\ref{fig:tsne}), for example through metric 
	learning as in our methods,  	
	 facilitates a better accuracy of the final classifier.
\end{itemize}

\subsubsection{Comparison with a transfer learning method}
We compared the pre-trained models of \textsc{GraphSage} and \textsc{GAT}
in~\cite{hlugzlplstrategies} and~\cite{lu2021learning} with \textsc{GraphSage} and \textsc{GAT} trained in \ourfinetune.
Specifically, we obtained the weights of \textsc{GraphSage} and \textsc{GAT} pretrained in~\cite{hlugzlplstrategies}, and they are referred to as \textsc{GraphSage (\transfer)} and \textsc{GAT (\transfer)}. We pre-trained \textsc{GraphSage} and \textsc{GAT} with the same architecture (i.e., $5$ layers, $300$ dimensional hidden units, and global mean pooling) by the code provided in \cite{lu2021learning} and denoted them as \textsc{GraphSage (\lp)} and \textsc{GAT (\lp)}.
We also pretrained \textsc{GraphSage} and \textsc{GAT} with the same architecture  according to the first training stage (see Section~\ref{subsec:firststage}), and they are referred to as \textsc{GraphSage (\ourfinetune)} and \textsc{GAT (\ourfinetune)}.
They were trained with different values of the margin hyperparameter $\alpha$ as specified in Section~\ref{subsubsec:hyperparameter}. In the classifier training step, \textsc{GraphSage (\transfer)}, \textsc{GAT (\transfer)}, \textsc{GraphSage (\lp)}, \textsc{GAT (\lp)}, \textsc{GraphSage (\ourfinetune)}, and \textsc{GAT (\ourfinetune)} were fine-tuned according to the second training stage (see Section~\ref{subsec:secondstage}) with the same search space of the architecture of the final classifier as specified in Section~\ref{subsubsec:hyperparameter}.
The other settings were the same as those described in Section~\ref{subsubsec:procedure}.

\begin{table*}[htbp!]
	\caption{\ourfinetune outperforms the 
	two transfer learning methods (\transfer \protect\cite{hlugzlplstrategies} and \lp \protect\cite{lu2021learning})
	in terms of classification accuracy in most cases when the 
	benchmark datasets are used.
	Note that \ourfinetune has several advantages over the  transfer learning methods, as highlighted in 
	Table~\ref{tab:compare}. The last column shows how much higher the average accuracy of \ourtriplet is than that of \transfer and \lp.
	}
	\label{tab:triplet_pretrain}
	\centering
	\scalebox{0.85}{\renewcommand{\arraystretch}{1.0}
		
		\begin{tabular}{@{}l||c|c|c|c|c|c||c@{}}
			\toprule
			\textbf{Method} & \multicolumn{6}{c}{\textbf{Dataset}}  & \textbf{Average Gap} \\ \cmidrule{2-7}	
			& {\textsc{DD}} & {\textsc{MUTAG}} & 
			{\textsc{MUTAG2}} & {\textsc{PTC-FM}} & 
			{\textsc{PROTEINS}} & {\textsc{IMDB-B}} & (in \% points)
			\\ \midrule

			\textsc{GAT (\ourfinetune)}  & \textbf{74.11 $\pm$ 1.42}
			& \textbf{78.09 $\pm$ 1.25}  & \textbf{76.62 $\pm$ 1.37}
			 & \textbf{61.56 $\pm$ 0.59} & \textbf{72.67 $\pm$ 0.67}  
			& \textbf{67.27	$\pm$ 0.81} & -
			\\     
			\textsc{GAT (\transfer)}  & 72.24 $\pm$ 0.83  &  
			76.86 $\pm$ 1.35 & 75.59 $\pm$ 1.48  & 61.28 $\pm$ 0.97
			& 71.76 $\pm$ 0.77 & 65.16	$\pm$ 1.47     & 1.56
			 \\    
			 \textsc{GAT (\lp)}  & 73.87 $\pm$ 0.95  &  
			 75.98 $\pm$ 0.92 & 76.38 $\pm$ 1.25  & 60.73 $\pm$ 1.02
			& 71.48 $\pm$ 0.93 & 66.59	$\pm$ 1.21     & 0.88
			 \\    
			 \midrule
			 \textsc{GraphSage (\ourfinetune)}  & \textbf{76.29 $\pm$ 1.25} & 
			 81.12 $\pm$ 0.61  &   \textbf{77.62 $\pm$ 0.47}  & 
			 \textbf{62.87 $\pm$ 0.63} & 72.51 $\pm$ 0.59 &
			 \textbf{68.18 $\pm$ 1.03}   & -
			 \\     
			 \textsc{GraphSage (\transfer)} & 75.26 $\pm$ 1.36 &  
			 \textbf{82.43 $\pm$ 1.49}   & 76.83 $\pm$ 0.95 &
			  62.61 $\pm$ 0.78 & 72.15 $\pm$ 0.83 &
			  67.14  $\pm$  0.52  
			 & 0.31
			 \\
			 \textsc{GraphSage (\lp)}  & 74.87 $\pm$ 1.13  &  
			81.56 $\pm$ 1.25 & 77.35 $\pm$ 0.64  & 61.13 $\pm$ 0.76
			& \textbf{72.78 $\pm$ 0.69} & 68.03	$\pm$ 0.78     & 0.48
			 \\    
			\bottomrule
	\end{tabular}}
\end{table*}

\begin{table*}[htbp!]
	\caption{
	\ourfinetune leads to higher classification accuracy than the two transfer learning methods (\transfer \protect\cite{hlugzlplstrategies} and \lp \protect\cite{lu2021learning}) in most NYC Taxi datasets.
	Compared with Table~\ref{tab:triplet_pretrain}, the gap is larger as 
	the 
	GNNs for transfer learning were pre-trained on a biology
	dataset, which is of a far domain. The last column shows how much higher the average accuracy of \ourtriplet is than that \transfer and \lp.}
	\label{tab:triplet_pretrain2}
	\centering
	\scalebox{0.85}{\renewcommand{\arraystretch}{1.0}
		
		\begin{tabular}{@{}l||c|c|c|c|c|c||c@{}}
			\toprule
			\textbf{Method} & \multicolumn{6}{c}{\textbf{Dataset}} & \textbf{Average Gap} \\ \cmidrule{2-7}	
			& {\textsc{Jan.G.}} & {\textsc{Feb. G.}} & 
			{\textsc{Mar. G.}} & {\textsc{Jan. Y.}} &
			 {\textsc{Feb. Y.}} & {\textsc{Mar. Y.}}  & (in \% points)
			\\ \midrule

			\textsc{GAT (\ourfinetune)} &  \textbf{74.47 $\pm$ 0.87} & 
			\textbf{68.19 $\pm$ 0.57}
			 &  \textbf{69.19 $\pm$ 1.02}
			  & \textbf{74.72 $\pm$ 1.23} & \textbf{68.81 $\pm$ 0.63} & 
			  \textbf{70.51 $\pm$ 0.93} & - 
			\\     
			\textsc{GAT (\transfer)} &  73.87 $\pm$ 1.13 & 65.89 $\pm$ 1.04 
		 & 67.25 $\pm$ 1.27
			  & 71.87 $\pm$ 1.35  & 66.24 $\pm$ 0.92 
			& 68.95 $\pm$ 1.25 & 1.97  \\
			\textsc{GAT (\lp)} & 73.36 $\pm$ 0.75 & 66.84 $\pm$ 0.97
		 & 67.91 $\pm$ 1.14
			  & 74.08 $\pm$ 1.16 & 67.59 $\pm$ 0.87
			& 68.37 $\pm$ 0.98 & 1.29 \\   
						\midrule
			\textsc{GraphSage (\ourfinetune)} & \textbf{76.41 $\pm$ 0.93}
			&   67.79 $\pm$ 0.82
			& \textbf{69.02 $\pm$ 1.25}  &  \textbf{75.33 $\pm$ 1.48}  &  67.61 
			$\pm$ 0.71
			& 70.21 $\pm$ 1.32 & - \\     
			\textsc{GraphSage (\transfer)} & 75.19 $\pm$ 0.98  & 67.82 $\pm$ 
			0.43 
			&   68.03 $\pm$ 0.66 & 73.66 $\pm$ 1.27  &  \textbf{68.24 $\pm$ 0.79} 
			& 70.25 $\pm$ 0.73 & 0.53  \\  
				\textsc{GraphSage (\lp)} & 75.04 $\pm$ 0.83 & \textbf{68.24 $\pm$ 0.76}	 & 68.37 $\pm$ 0.96  & 74.17 $\pm$ 1.16 &
				67.08 $\pm$ 0.54  & \textbf{70.96 $\pm$ 0.82} & 0.42 \\   
			\bottomrule
	\end{tabular}}
\end{table*}

As shown in Tables~\ref{tab:triplet_pretrain} and~\ref{tab:triplet_pretrain2}, despite being pre-trained on a much smaller dataset, \ourfinetune achieves better accuracy in $83\%$ of the considered cases: up to $2\%$ 
points in  the benchmark 
datasets and up to $3\%$ points in the NYC Taxi 
datasets. This validates our claims in Table~\ref{tab:compare}.

\subsubsection{Running time}
For each hyperparameter setting in a dataset, the original setting takes up to 
half an hour to train a model.  Due to having two stages, \ourtriplet and 
\ourfinetune take up to an hour for both stages. 
In~\cite{hlugzlplstrategies}, the transfer-learning method was reported to 
take up to one day to pre-train on a rich dataset. Pre-training each GNN architecture accordingly to~\cite{lu2021learning} took several hours to complete.

\subsection{Further analysis}
We study the final embeddings, generated by three training methods: original, \ourtriplet, \ourfinetune. In Fig.~\ref{fig:dd}, despite having 2 dimensions, the embeddings of the original setting can be reduced to 1 dimension. We suspect that given the potential capacity, specifically the dimension of the final embeddings, \ourtriplet and \ourfinetune utilize such expressive power better than the original setting by resulting in higher intrinsic dimension, weaker correlation between dimensions, and larger distance between embeddings~\cite{chen2020measuring}. While measuring the distance did not provide clear evidence, the results of the other two measurements validate our hypothesis.

\subsubsection{Intrinsic dimension}
\begin{figure*}[t!]
	\centering
	{
	\begin{subfigure}{0.325\textwidth}
	    \centering
		\includegraphics[width=1.72in]{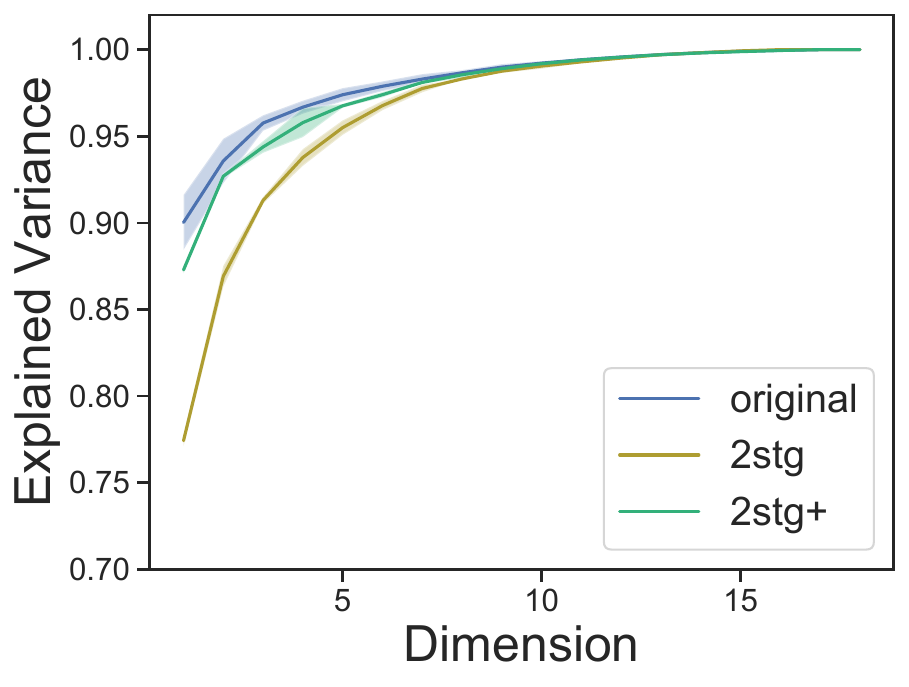}
		\caption{\textsc{GraphSage}}
	\end{subfigure}
	\begin{subfigure}{0.325\textwidth}
	    \centering
		\includegraphics[width=1.72in]{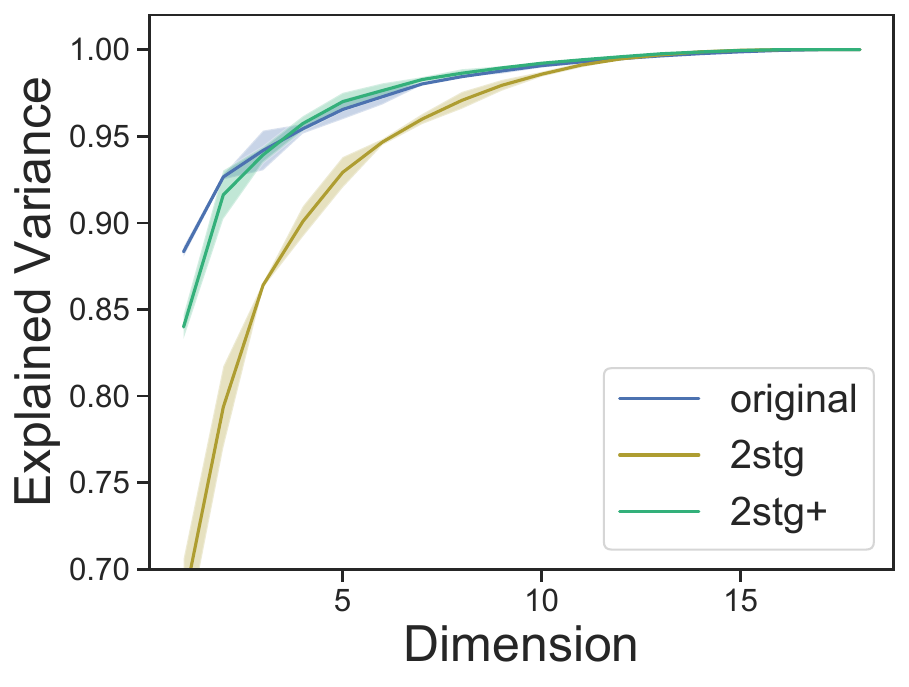}
		\caption{\textsc{GAT}}
	\end{subfigure}
	\begin{subfigure}{0.325\textwidth}
	    \centering
		\includegraphics[width=1.72in]{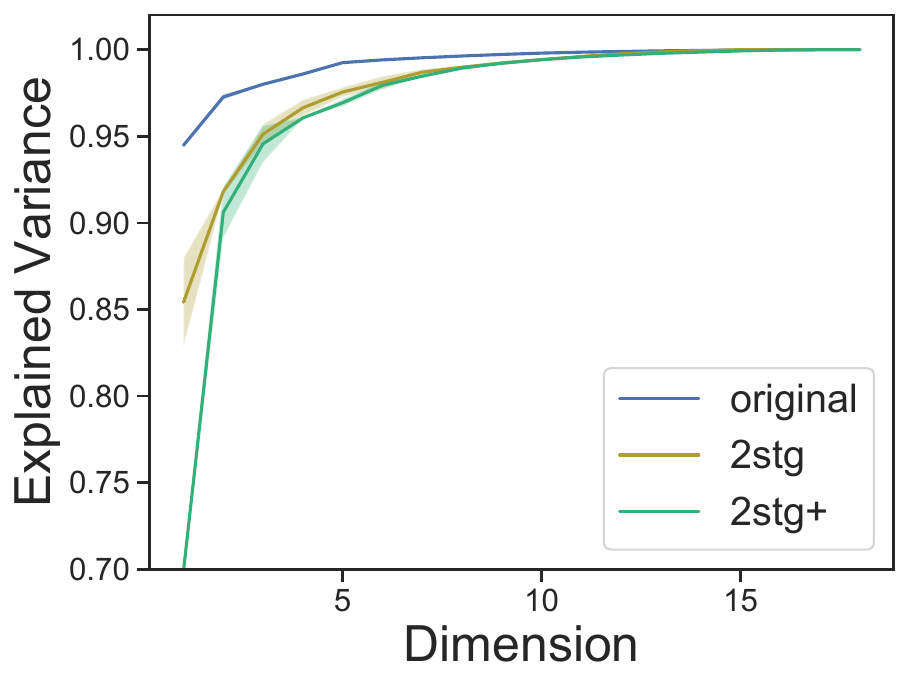}
		\caption{\textsc{DiffPool}}
	\end{subfigure}\\
	}
	\centering{
	\begin{subfigure}{0.325\textwidth}
	    \centering
		\includegraphics[width=1.72in]{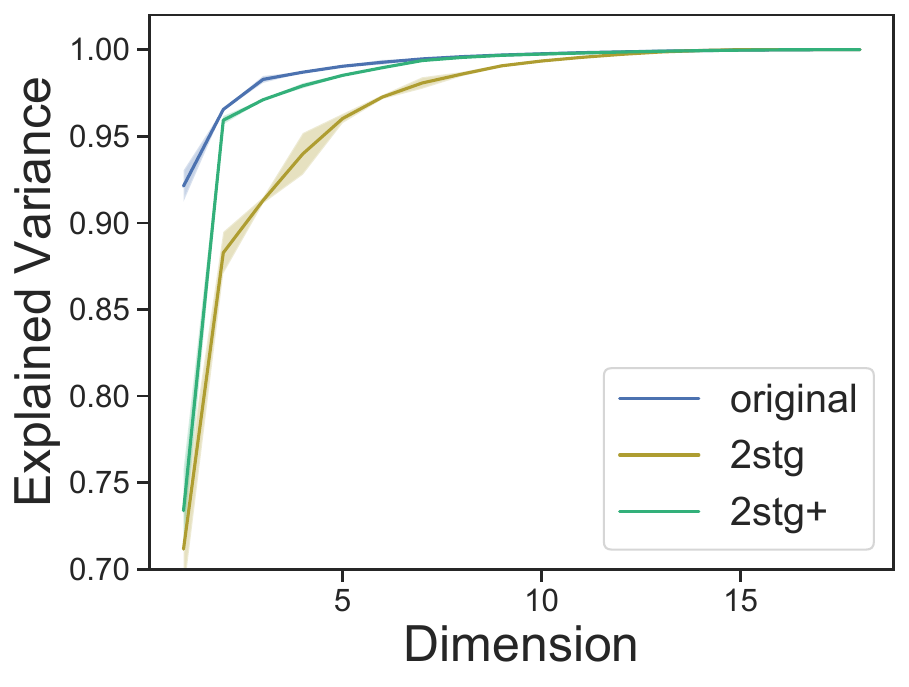}
		\caption{\textsc{EigenGCN}}
	\end{subfigure}
	\begin{subfigure}{0.325\textwidth}
	    \centering
		\includegraphics[width=1.72in]{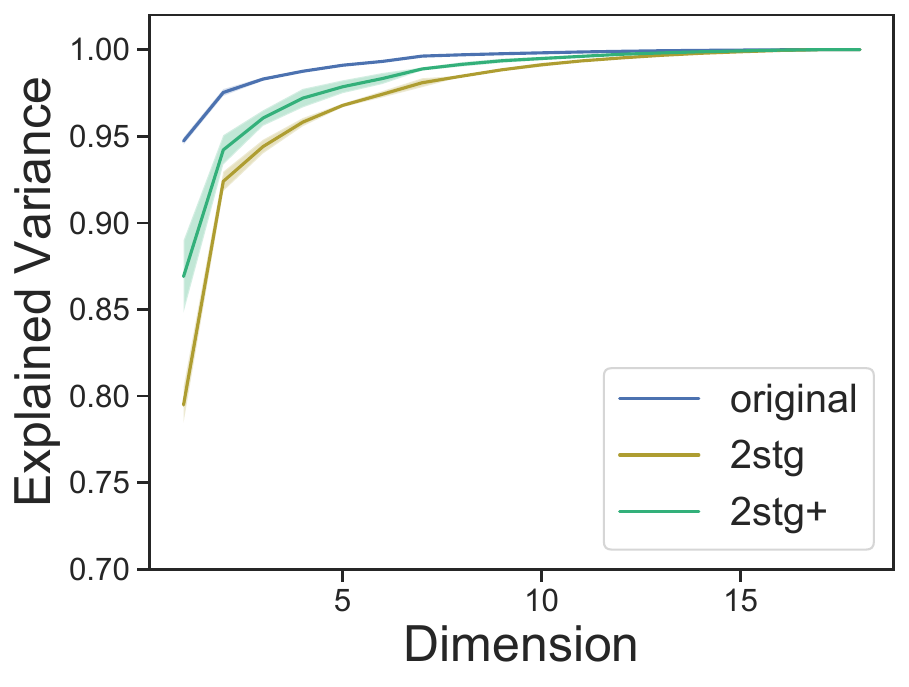}
		\caption{\textsc{SAGPool}}
	\end{subfigure}
	}
		
	\caption{\label{fig:pca_mutag}
		Explained variance with respect to dimensions when applying 
		PCA to the embeddings generated by each GNN trained in the original setting, 
		\ourtriplet, and \ourfinetune on the \textsc{MUTAG} dataset. Means and standard deviation of 5 runs are plotted. The 
		respective intrinsic dimensions are reported in 
		Table~\ref{tab:eff_dim}. In order to 
		retain $99\%$ 
		of the 
		explained variance, the embeddings obtained in the original setting 
		require lower dimensions than those obtained in \ourtriplet 
		and \ourfinetune.
	}
\end{figure*}

We measure the \textit{intrinsic dimension}, which we define as the dimension 
needed to retain 99$\%$ of the variance in the final embeddings. In the final 
embeddings of dimension $d$ of the validation set, we apply Principal Component 
Analysis (PCA) and measure the cumulative explained variance $V(i)$ achieved by 
setting the final dimension as $i$ for $i=1,2,..,d$. Assume that $V(j) \leq 
0.99 \leq V(j+1)$, we linearly interpolate between $j$ and $(j+1)$ to obtain 
the dimension with $99\%$ explained variance. Such dimensions are reported in 
Tables~\ref{tab:eff_dim} and~\ref{tab:eff_dim2}. Given $64$ dimensions as the 
capacity, \ourtriplet 
and \ourfinetune retain a higher intrinsic dimension, implying that they 
utilize this capacity better than the original setting. As an 
illustrative 
example from the \textsc{MUTAG} dataset in Fig.~\ref{fig:pca_mutag} shows, most of the 
variance can be explained by lower dimensions in the original 
setting than in \ourtriplet and \ourfinetune.

\begin{table*} [htbp]
	\caption{Intrinsic dimension obtained by PCA on embeddings of dimension 64 for the benchmark datasets. Each entry is the mean of 5 runs. In most cases, the GNNs trained in \ourtriplet and \ourfinetune lead to higher intrinsic dimensions than that trained in the original setting.}
	\label{tab:eff_dim}
	\centering
	\scalebox{0.85}{
		
		\begin{tabular}{@{}l||A|A|A|A|A|B
		 @{}}
			\toprule
			\textbf{Method} & \multicolumn{6}{c}{\textbf{Dataset}} \\ 
		\cmidrule{2-7}
			& {\textsc{DD}} & 
			{\textsc{MUTAG}} & 
			{\textsc{MUTAG2}}
			& {\textsc{PTC-FM}} & 
			{\textsc{PROTEINS}} & 
			{\textsc{IMDB-B}}
			\\ \cmidrule{1-7}
			\textsc{GraphSage}  & 37.01  &  9.03 & 10.23  &  8.76
			  & 33.67  & \textbf{24.41} \\ 
			\textsc{GraphSage (\ourtriplet)} & \textbf{38.83} & \textbf{9.98} & 17.21 & \textbf{19.98}
			&  \textbf{36.13}  &  24.18 	\\     
			\textsc{GraphSage (\ourfinetune)} & 18.33  & 9.18 &  \textbf{30.27} & 16.65
			  & 18.11  &  14.31 \\     
			\cmidrule{1-7}
			\textsc{GAT}  & 8.38  & 9.65  & 12.26  &  11.07
			 &  33.72 & \textbf{26.46}  \\ 
			\textsc{GAT (\ourtriplet)} &  \textbf{34.53}  &  \textbf{10.88}  &  \textbf{20.47}  & \textbf{18.73}
			  &  31.13 &  15.33 \\     
			\textsc{GAT (\ourfinetune)} & 31.17  & 9.37	 & 18.54   & 16.94 
			  & \textbf{33.88}  &  14.54 \\     
			\cmidrule{1-7}
			\textsc{DiffPool}  & 6.69  & 3.69  &  15.97 & 13.17
			  & 14.33  &  14.31 \\ 
			\textsc{DiffPool (\ourtriplet)}  & \textbf{37.16}  & 7.94 & 22.64  & \textbf{20.02}
			  &  \textbf{18.25} &	 \textbf{19.39} \\     
			\textsc{DiffPool (\ourfinetune)} &  15.11 & \textbf{8.08} & \textbf{25.19} & 7.02
			  & 10.21  & 11.21 \\     
			\cmidrule{1-7}
			\textsc{EigenGCN}  & 14.76 & 4.98 & 20.49 & 12.57
			  &  17.94 &	17.21  \\ 
			\textsc{EigenGCN (\ourtriplet)}  & \textbf{25.63} & \textbf{8.71} & \textbf{31.53} & \textbf{19.77}
			  &  19.26 &	\textbf{17.32}  	\\     
			\textsc{EigenGCN (\ourfinetune)} & 22.76 & 5.38 &  28.75 & 19.62
			  &  \textbf{19.57} &  11.74 \\     
			\cmidrule{1-7}
			\textsc{SAGPool}  & 15.86 & 2.81  &17.88   & 12.99 
			  & 11.82 & 17.31  \\ 
			\textsc{SAGPool (\ourtriplet)}  & \textbf{29.46} & \textbf{9.76} & 18.25 & \textbf{15.67}
			  & \textbf{16.83} & 	\textbf{18.49}  \\     
			\textsc{SAGPool (\ourfinetune)} & 25.83 & 7.31 & \textbf{18.37} & 14.74
			  & 15.87 &	 11.75 \\    
			\bottomrule
	\end{tabular}}
\end{table*}

\begin{table*}[t!]
	\caption{Intrinsic dimension obtained by PCA on embeddings of dimension 64 for the New York Taxi datasets. Each entry is the mean of 5 runs. In almost all cases, the embeddings obtained by the GNNs trained in \ourtriplet and \ourfinetune have higher intrinsic dimensions than those obtained by the GNNs trained in the original setting.}
	\label{tab:eff_dim2}
	\centering
	\scalebox{0.85}{ 
		
		\begin{tabular}{@{}l||A|A|A|A|A|B@{}}
			\toprule
			\textbf{Method} & \multicolumn{6}{c}{\textbf{Dataset}} \\ 
		\cmidrule{2-7}
			
			& {\textsc{Jan. G.}} & {\textsc{Feb. G.}} & {\textsc{Mar. G.}}
			& {\textsc{Jan. Y.}} & {\textsc{Feb. Y.}} & {\textsc{Mar. Y.}} 
			\\ \cmidrule{1-7}
			\textsc{GraphSage}  & 26.46 &  30.49 & 32.21  & 31.02  &  21.24 & 23.41\\ 
			\textsc{GraphSage (\ourtriplet)} & 35.98  & \textbf{36.03}  &   \textbf{36.98} &  \textbf{38.95}  &  \textbf{38.61}   & \textbf{36.64}	\\     
			\textsc{GraphSage (\ourfinetune)}  &  \textbf{37.56} &  33.37  & 29.88   & 24.06  &  23.51 &	32.07\\     
			\cmidrule{1-7}
			\textsc{GAT}    &  35.17 &  34.44  & 33.61   & 32.01  &  \textbf{36.95} & \textbf{31.72}\\ 
			\textsc{GAT (\ourtriplet)}  &  23.57 &  25.24  &  \textbf{38.01} & \textbf{39.02}   & 24.34  &	23.33\\     
			\textsc{GAT (\ourfinetune)}  &  \textbf{36.89} &  \textbf{38.27}  & 26.98  & 35.18   &  23.57 &	28.42\\     
			\cmidrule{1-7}
			\textsc{DiffPool}  &  4.64 &  16.84  & 6.62  & 16.56  &  16.19 & 15.71\\ 
			\textsc{DiffPool (\ourtriplet)}  &  \textbf{33.03} & 32.77   & \textbf{26.78}  & \textbf{36.01}  & \textbf{33.09}  & \textbf{33.93}\\     
			\textsc{DiffPool (\ourfinetune)} & 32.57  & \textbf{33.14}   & 26.22  & 30.93  & 20.84  &	30.35\\     
			\cmidrule{1-7}
			\textsc{EigenGCN}  & 3.98  & 15.59   & 3.19  & 15.77  & 28.02  & 23.47\\ 
			\textsc{EigenGCN (\ourtriplet)}  & 31.21  & 32.45   & \textbf{27.68}  & \textbf{32.81}  & 26.55   & \textbf{26.92}	\\     
			\textsc{EigenGCN (\ourfinetune)}  & \textbf{32.26}  &  \textbf{33.98}  & 18.87  & 30.87  & \textbf{31.01}   &	26.47\\     
			\cmidrule{1-7}
			\textsc{SAGPool}   & 5.42  & 17.66   & 3.75  & 14.99  & 25.09   & 20.01\\ 
			\textsc{SAGPool (\ourtriplet)}   & 31.81  & 28.57   & \textbf{31.19}  & \textbf{32.71}  &  \textbf{28.19}   &	\textbf{31.11}\\     
			\textsc{SAGPool (\ourfinetune)} & \textbf{33.18}  &  \textbf{32.49}  & 28.16  &21.31   &  27.56   &	29.11\\    
			\bottomrule
	\end{tabular}
}
\end{table*}

\subsubsection{Correlation between final embedding dimensions}
\begin{figure*}[t]
	\centering{
	\begin{subfigure}{0.325\textwidth}
	    \centering
		\includegraphics[width=1.72in]{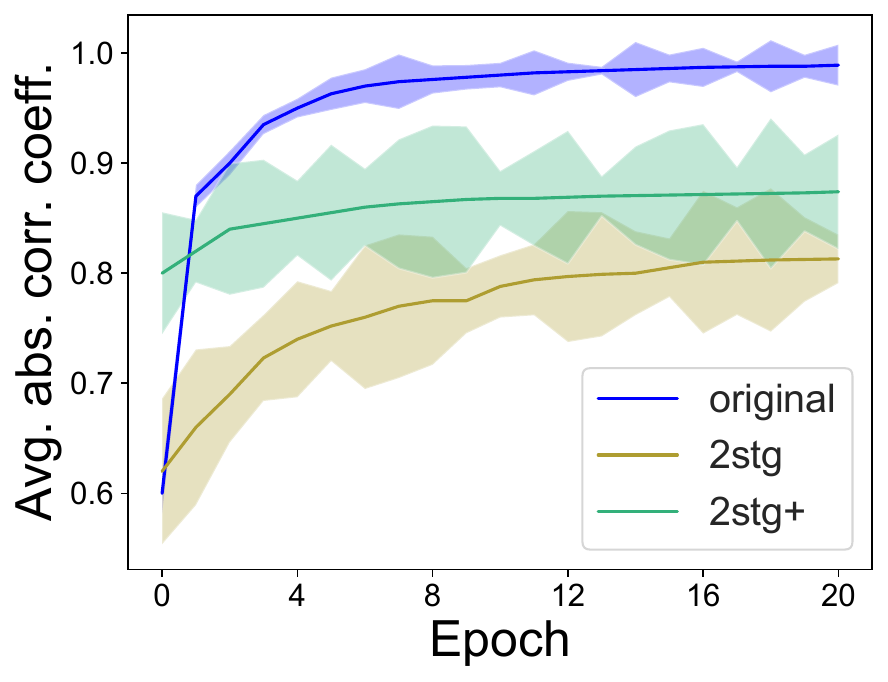}
		\caption{\textsc{GraphSage}}
	\end{subfigure}
	\begin{subfigure}{0.325\textwidth}
	    \centering
		\includegraphics[width=1.72in]{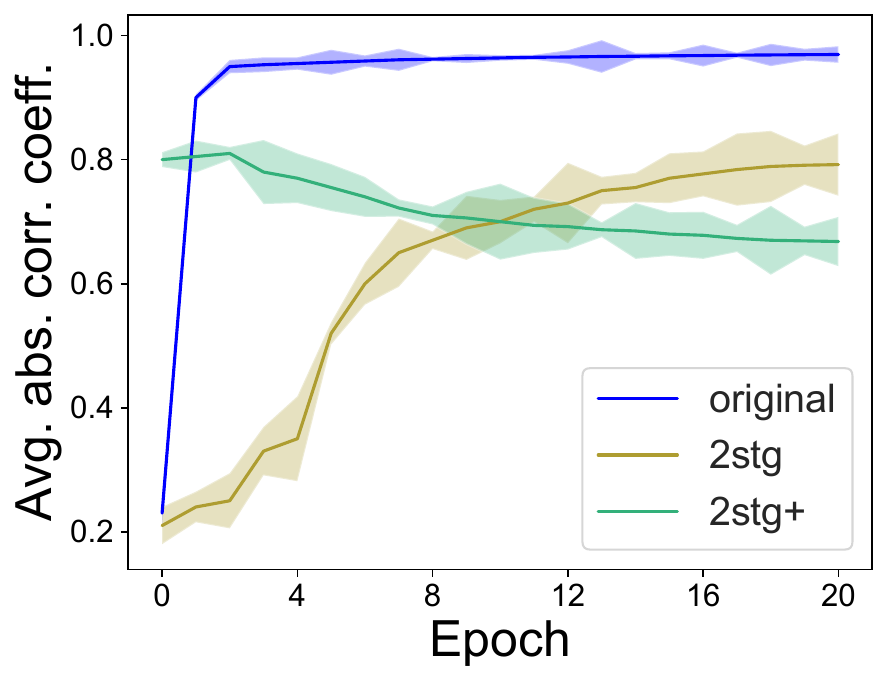}
		\caption{\textsc{GAT}}
	\end{subfigure}
	\begin{subfigure}{0.325\textwidth}
	    \centering
	    \includegraphics[width=1.72in]{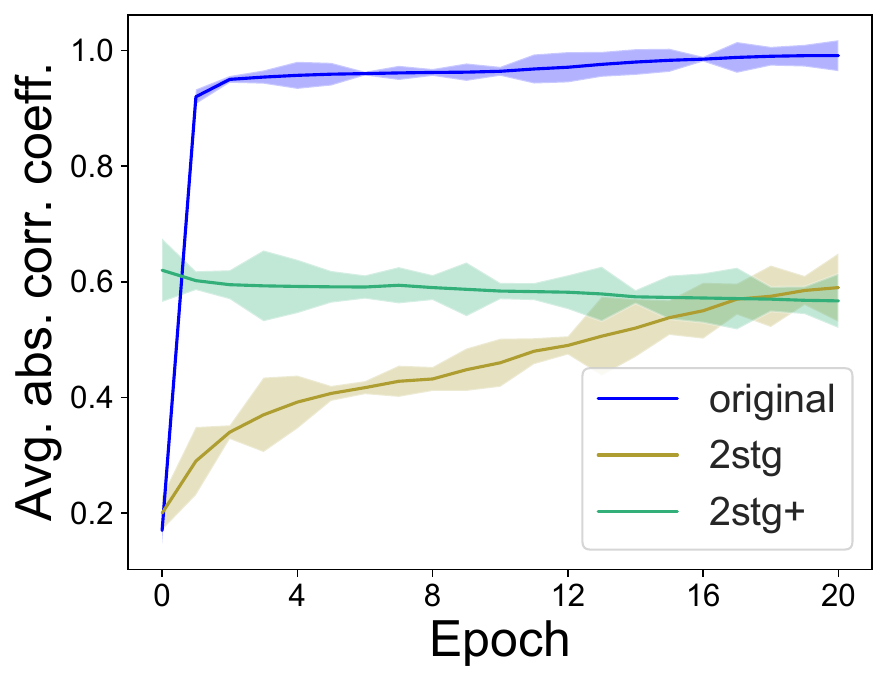}
	    \caption{\textsc{DiffPool}}
	\end{subfigure}\\
	}
	\centering
	{
	\begin{subfigure}{0.325\textwidth}
	\centering
	\includegraphics[width=1.72in]{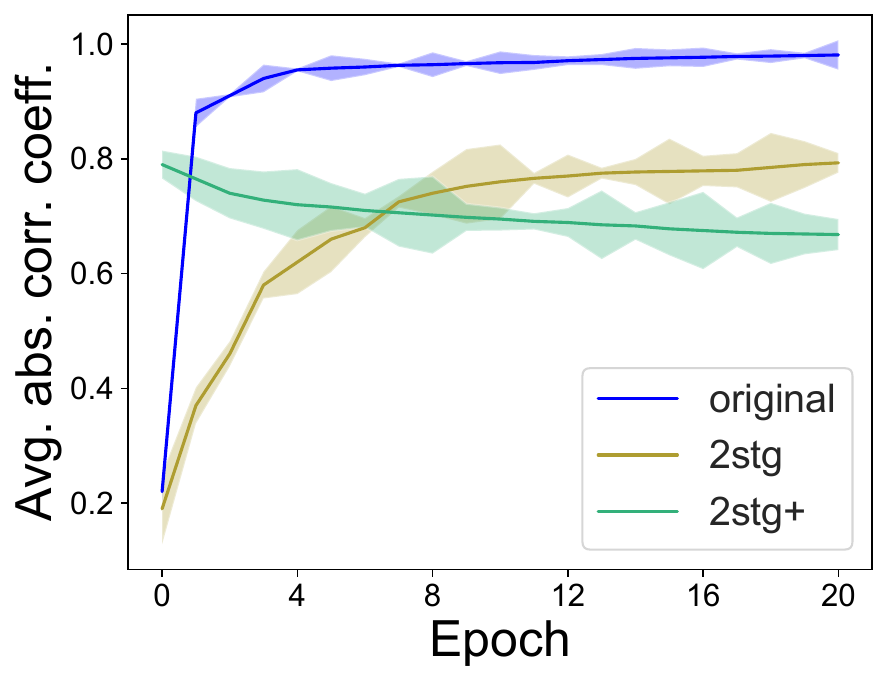}
	\caption{\textsc{EigenGCN}}
	\end{subfigure}
	\begin{subfigure}{0.325\textwidth}
	\centering
	\includegraphics[width=1.72in]{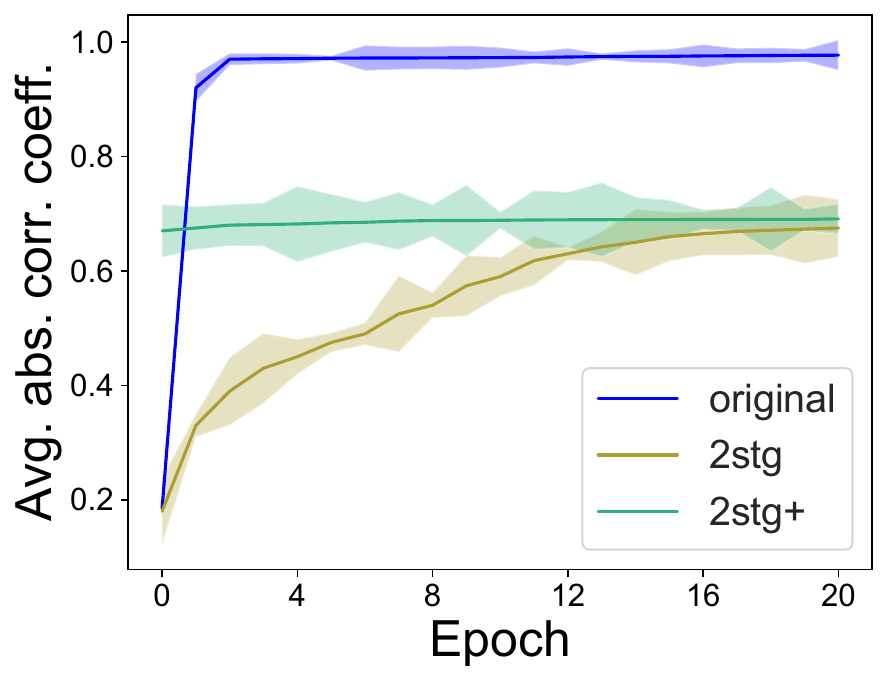}
	\caption{\textsc{SAGPool}}
	\end{subfigure}
	}
	\caption{\label{fig:green1}
		Average absolute correlation coefficients over epochs on the Jan. G. dataset. Means and standard deviations of 15 runs are plotted. Embedding dimensions quickly become strongly correlated in the orginal setting. Such correlation is much weaker in \ourtriplet and \ourfinetune.
	}
\end{figure*}

We measure the average absolute correlation coefficients between final embedding dimensions. Specifically, given final embeddings of dimension five, the 
average of absolute values of ${5 \choose 2}$ pairwise correlation coefficients is 
calculated. As seeen in Tables~\ref{tab:pair_corr1} and~\ref{tab:pair_corr2}, 
in most of the cases, the original setting leads to stronger 
correlation than \ourtriplet and \ourfinetune, and several obtained dimensions
are too strongly correlated, making them redundant. This implies that 
\ourtriplet and \ourfinetune utilize the provided capacity (i.e., dimension of final embeddings) better than the original setting by giving fewer redundant dimensions.
As seen in Fig.~\ref{fig:green1}, the average absolute correlation coefficients in the original setting quickly become close to 1. In \ourtriplet and \ourfinetune, such correlation is much weaker.

\begin{table*} [htbp]
	\caption{Average absolute correlation coefficients between the five 
	dimensions of the final embeddings for the benchmark datasets. Each entry 
	is the mean of 5 runs, and it ranges from $0$ to $1$. Higher values 
	indicate stronger correlation between the dimensions. In all cases, the 
	original setting leads to stronger correlation between the dimensions than 
	\ourtriplet and \ourfinetune.}
	\label{tab:pair_corr1}
	\centering
	\scalebox{0.83}{ \renewcommand{\arraystretch}{1.0}
		
		\begin{tabular}{@{}l||A|A|A|A|A|B @{}}
			\toprule
			\textbf{Method} & \multicolumn{6}{c}{\textbf{Dataset}} \\ 
		\cmidrule{2-7}
			& {\textsc{DD}} & {\textsc{MUTAG}} & {\textsc{MUTAG2}}
			& {\textsc{PTC-FM}} & {\textsc{PROTEINS}} & {\textsc{IMDB-B}}
			\\ \cmidrule{1-7}
			\textsc{GraphSage}  & \textbf{0.94}  &  \textbf{0.77} & \textbf{0.91}  &  \textbf{0.56} 	  & \textbf{0.96}  & \textbf{0.55}
			\\ 
			\textsc{GraphSage (\ourtriplet)} & 0.67 &  0.66 & 0.26 & 0.28
			&  0.43  &  0.31 	\\     
			\textsc{GraphSage (\ourfinetune)} & 0.62  & 0.59  & 0.42  & 0.45 
			  &  0.41 & 0.32 	\\     
			\cmidrule{1-7}
			\textsc{GAT}  & \textbf{0.92}  &  \textbf{0.98} &  \textbf{0.87} &  \textbf{0.62}
			 &  \textbf{0.82} &   \textbf{0.84} \\ 
			\textsc{GAT (\ourtriplet)} & 0.72   &  0.45  &   0.25 & 0.26
			  & 0.46  & 0.53  \\     
			\textsc{GAT (\ourfinetune)} & 0.64  & 0.56	 & 0.29   & 0.37 
			  &  0.41 & 0.47 	\\     
			\cmidrule{1-7}
			\textsc{DiffPool}  & \textbf{0.81}  &  \textbf{0.85} & \textbf{0.75}  & \textbf{0.49}
			  &  \textbf{0.91} &  \textbf{0.87} \\ 
			\textsc{DiffPool (\ourtriplet)}  &  0.36 & 0.52 & 0.51 & 0.35
			  &  0.38 &	 0.39 \\     
			\textsc{DiffPool (\ourfinetune)} & 0.32  & 0.44 & 0.47 & 0.31
			  &  0.42 & 0.33  \\     
			\cmidrule{1-7}
			\textsc{EigenGCN}  & \textbf{0.72} & \textbf{0.92} & \textbf{0.71} & \textbf{0.58}
			  & \textbf{0.84}  &	\textbf{0.78} \\ 
			\textsc{EigenGCN (\ourtriplet)}  & 0.58 & 0.51 & 0.47 & 0.45
			  & 0.57  &	 0.34 	\\     
			\textsc{EigenGCN (\ourfinetune)} & 0.54 & 0.75 & 0.42 & 0.37
			  &  0.43 &  0.32 \\     
			\cmidrule{1-7}
			\textsc{SAGPool}  & \textbf{0.75}  & \textbf{0.98}  & \textbf{0.62}  & \textbf{0.47}
			  &  \textbf{0.82} &   \textbf{0.82}\\ 
			\textsc{SAGPool (\ourtriplet)}  & 0.49 & 0.61 & 0.38 & 0.26
			  & 0.34 & 	0.44 	\\     
			\textsc{SAGPool (\ourfinetune)} & 0.41 & 0.55  & 0.36 & 0.31
			  & 0.32 &	 0.41 \\    
			\bottomrule
	\end{tabular}
}
\end{table*}

\begin{table*}[t!]
	\caption{Average absolute correlation coefficients between the five 
	dimensions of the final embeddings for the New York Taxi datasets. Each 
	entry is the mean of 5 runs, and it ranges from $0$ to $1$. Higher values 
	indicate stronger correlation between the dimensions. In most cases, the 
	original setting leads to stronger correlation between the dimensions than 
	\ourtriplet and \ourfinetune.}
	\label{tab:pair_corr2}
	\centering
	\scalebox{0.83}{ \renewcommand{\arraystretch}{1.0}
		
		\begin{tabular}{@{}l||A|A|A|A|A|B @{}}
			\toprule
			\textbf{Method} & \multicolumn{6}{c}{\textbf{Dataset}} \\ 
		\cmidrule{2-7}
		
			& {\textsc{Jan. G.}} & {\textsc{Feb. G.}} & {\textsc{Mar. G.}}
			& {\textsc{Jan. Y.}} & {\textsc{Feb. Y.}} & {\textsc{Mar. Y.}} 
			\\ \cmidrule{1-7}
			\textsc{GraphSage}   & \textbf{0.92} & \textbf{0.87}  & \textbf{0.95}  &  \textbf{0.89} & \textbf{0.89}  & \textbf{0.94}\\ 
			\textsc{GraphSage (\ourtriplet)}  & 0.41  & 0.66 & 0.66   & 0.68   & 0.61    & 0.84	\\     
			\textsc{GraphSage (\ourfinetune)} & 0.45  & 0.76   &  0.66  & 0.62  &  0.54 & 0.49	\\     
			\cmidrule{1-7}
			\textsc{GAT}   & \textbf{0.71}  & 0.77   & \textbf{0.91}   & \textbf{0.86}  & \textbf{0.78}  & \textbf{0.86}\\ 
			\textsc{GAT (\ourtriplet)} & 0.54  &  0.54  &  0.61 &  0.41  &  0.52 & 0.52\\     
			\textsc{GAT (\ourfinetune)}  & 0.65  &  \textbf{0.84}  &  0.58 &  0.48  &  0.58 & 0.46	\\     
			\cmidrule{1-7}
			\textsc{DiffPool}  & \textbf{0.95}  &  0.88  & \textbf{0.97}  & \textbf{0.98}  & 0.59  & \textbf{0.97}\\ 
			\textsc{DiffPool (\ourtriplet)}  & 0.66  & 0.51   &  0.61 & 0.76  & \textbf{0.83}  & 0.88\\     
			\textsc{DiffPool (\ourfinetune)}  & 0.57  &  \textbf{0.92}  & 0.59  & 0.44  & 0.39  &	0.53\\     
			\cmidrule{1-7}
			\textsc{EigenGCN}   &  \textbf{0.97} & \textbf{0.97}   & \textbf{0.92}  & \textbf{0.95}  & \textbf{0.94}  & \textbf{0.98}\\ 
			\textsc{EigenGCN (\ourtriplet)} &  0.68 & 0.78   & 0.56  & 0.44  & 0.53   &0.84	\\     
			\textsc{EigenGCN (\ourfinetune)} & 0.45  & 0.71   &  0.51 &  0.44 &  0.48  &	0.63\\     
			\cmidrule{1-7}
			\textsc{SAGPool}   &  \textbf{0.99} & \textbf{0.94}   & \textbf{0.88}  & 0.73  & \textbf{0.93}   & \textbf{0.92}\\ 
			\textsc{SAGPool (\ourtriplet)}   &   0.31 & 0.42   & 0.51  & 0.56  & 0.62    & 0.49	\\     
			\textsc{SAGPool (\ourfinetune)} & 0.38  &  0.41  &  0.57 & \textbf{0.92}  &    0.57 &	0.53\\    
			\bottomrule
	\end{tabular}
}
\end{table*}

\section{Conclusion}
\label{sec:conclusion}
Graph neural networks are powerful tools in dealing with the graph classification task. However, training them end-to-end to predict class probabilities often fails to 
realize their full capability.
Thus, we 
apply GNN models in a triplet framework to learn discriminative 
embeddings first, and then train a classifier on those embeddings. 
Extensive experiments in 12 datasets lead to following observations:
\begin{itemize}
	\item End-to-end training often fails to realize the full potential of GNN 
	models. Our method consistently improves the accuracy of $5$ (out of 
	$5$ tested) GNN models
	in $12$ (out of $12$ considered) datasets over the original setting.
	\item Despite not requiring any additional massive datasets or long training 
	time, our two-stage method obtains better accuracy than a 
	state-of-the-art pre-training method based on 
	transfer-learning in 83\% of the cases.	
	\item Training each GNN in our method utilizes the given capacity better by producing embeddings with higher intrinsic dimension and weaker correlation between dimensions.

\end{itemize}
As future work, we would like to address some current limitations of our method and expand the scope of our work. One possible approach is to reduce the overhead in triplet sampling to reduce the overall training time by using fast nearest-neighbor search techniques (e.g.,~\cite{park2015a,hajebi2011fast}). We will also consider employing the global loss~\cite{ktena2018metric} or the Siamese Community-Preserving loss objective~\cite{liu2019community} instead of the triplet loss. In addition, we will apply the two-stage training idea to other downstream tasks, such as node classification and edge label prediction.



\subsection*{Availability of data and code}
The source code and datasets used in the study are available at 
\url{http://github.com/manhtuando97/two-stage-gnn}. 

\appendix
\section{Appendix: Effect of Classification Layers}
\label{appendix:hyperparameter}

\begin{figure}[t!]
	\addtolength{\tabcolsep}{-3pt}
	\begin{center}
	    \textbf{\textsc{PTC-FM} Dataset:} \hfill \ \ \\
		\begin{subfigure}{0.45\textwidth}
		    \centering
			\includegraphics[width=2.3in]{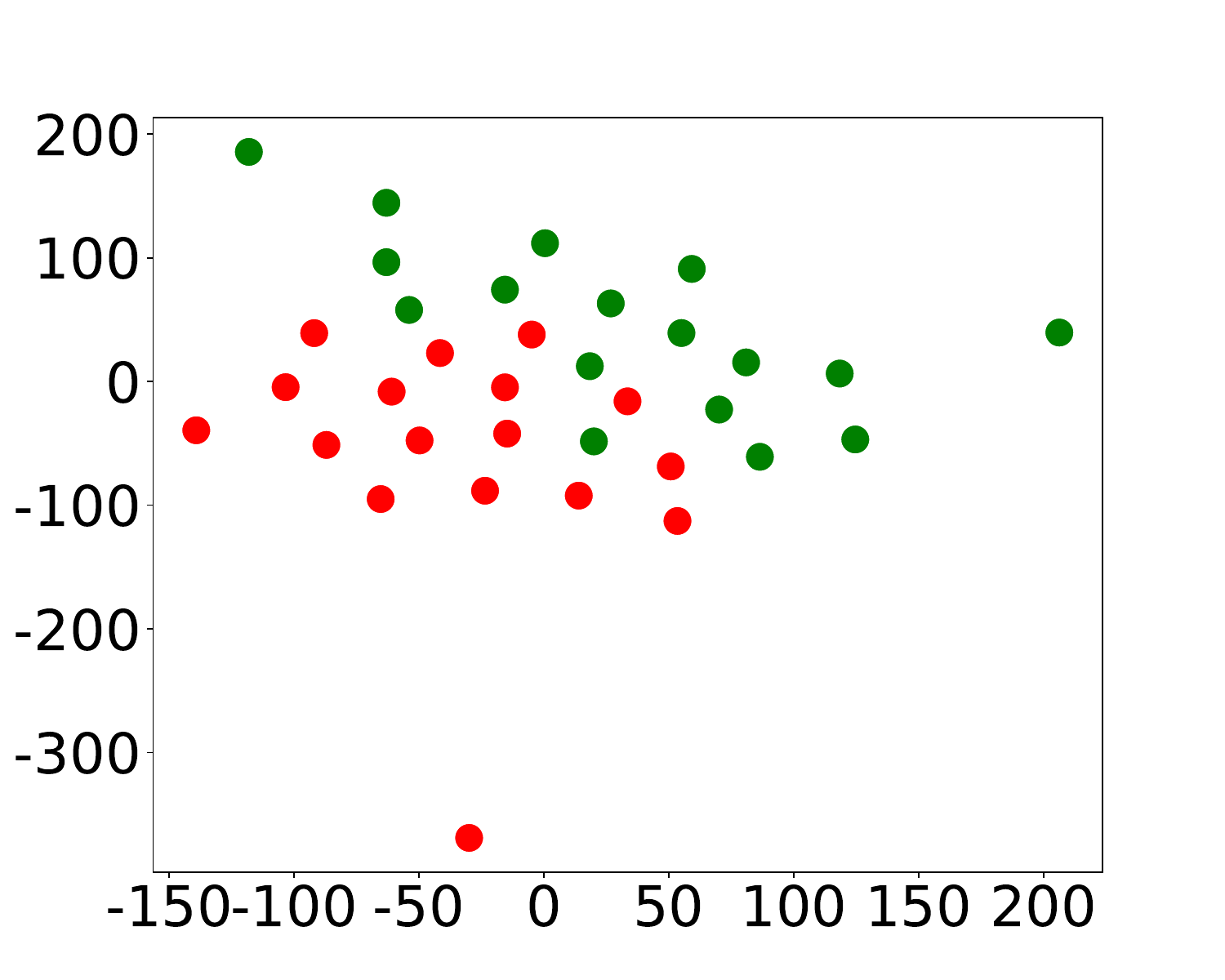} 
		\end{subfigure}
		\hspace{5mm}
		\begin{subfigure}{0.45\textwidth}
		    \centering
			\includegraphics[width=2.3in]{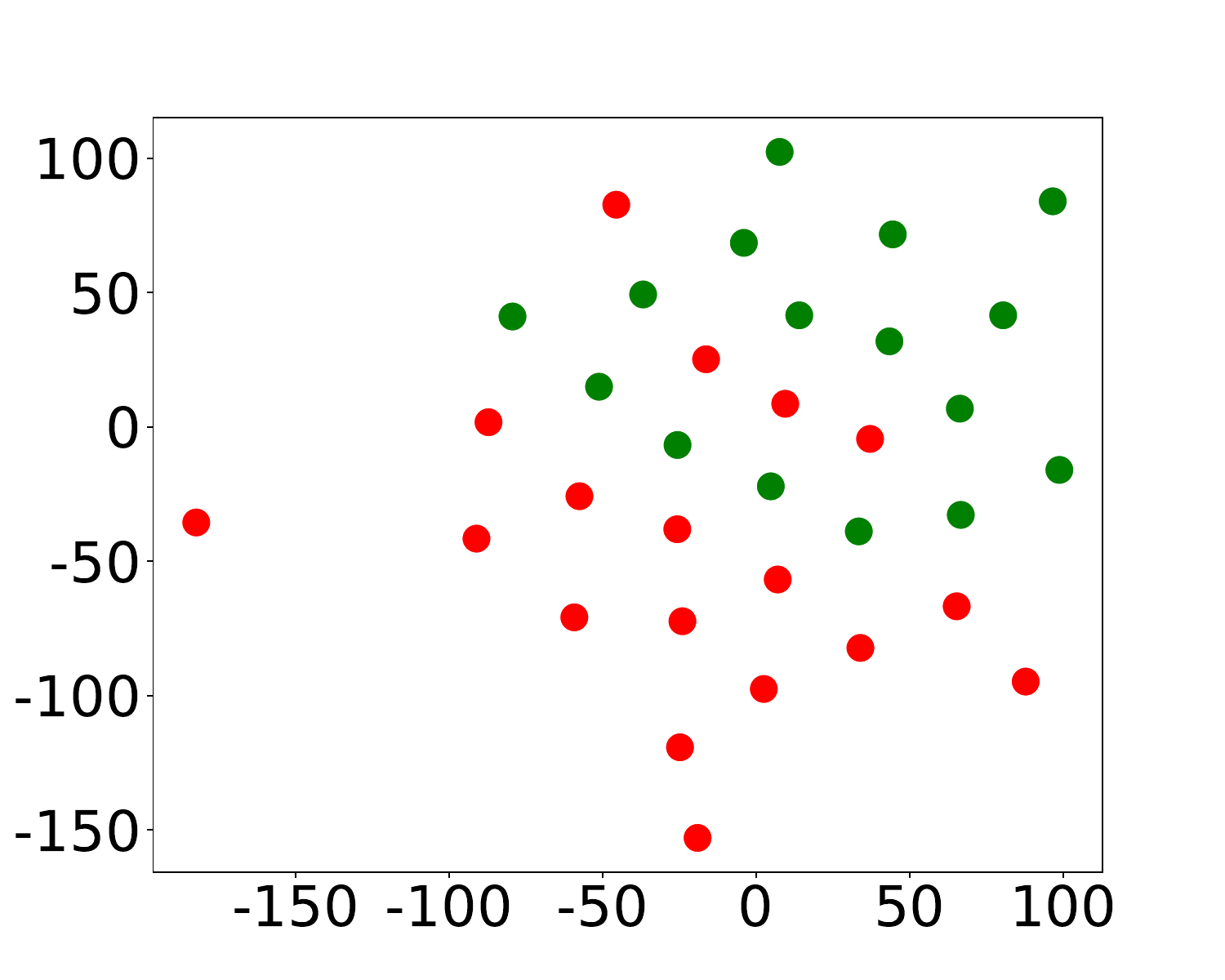}
		\end{subfigure} \\
		
		\vspace{1mm}
		\textbf{\textsc{Jan. G.} Dataset: } \hfill \ \ \\
		\begin{subfigure}{0.45\textwidth} 
		    \centering
			\includegraphics[width=2.3in]{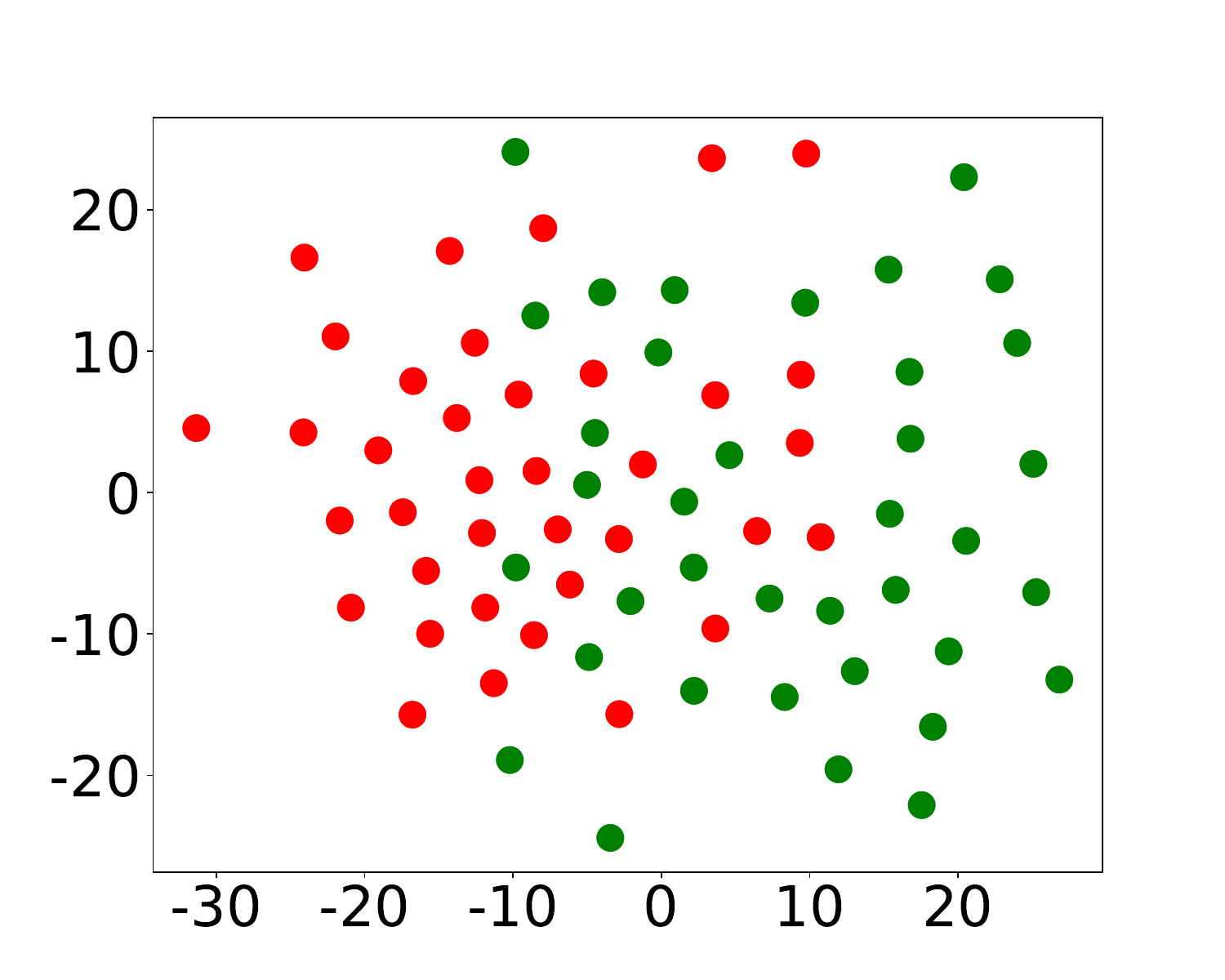}
			\caption{\ourfinetune with one classification layer}
		\end{subfigure}
			\hspace{5mm}
		\begin{subfigure}{0.45\textwidth}
		    \centering
			\includegraphics[width=2.3in]{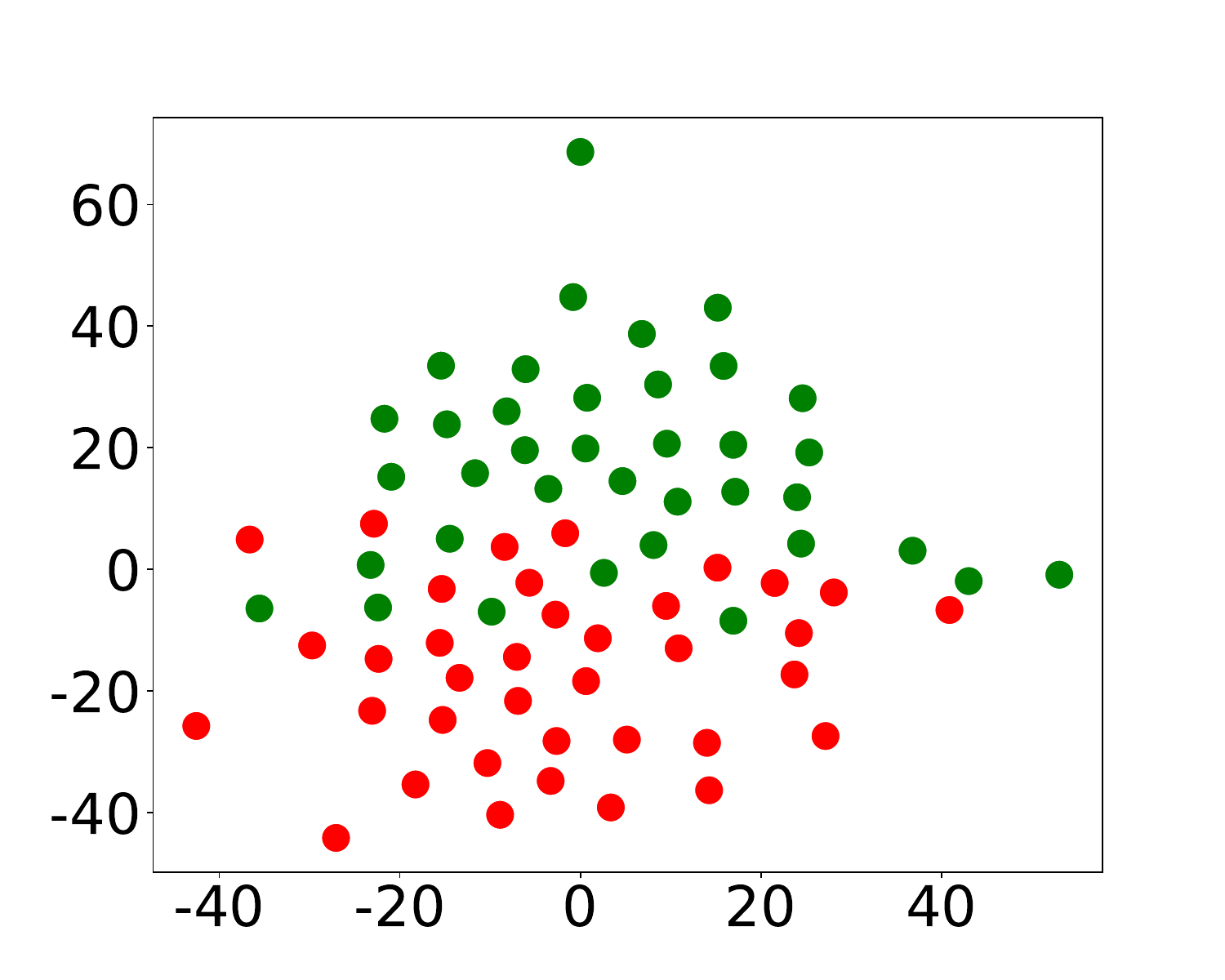}
			\caption{\ourfinetune with up to three classification layer}
		\end{subfigure}
		
		\caption{\label{fig:tsne2}
			Graph embeddings generated by \textsc{GraphSAGE} for the datasets \textsc{PTC-FM} and \textsc{JAN. G.} in the two cases of \ourfinetune with one classification layer (left) and \ourfinetune with up to three classification layers (right), respectively. These are the graphs embeddings (i.e., $\hat{f}(g_i)$ of each graph $g_i$), which are the input of the classification part in Fig.~\ref{fig:stage2}. T-SNE is applied to reduce the embedding dimension from 64 to 2 for visualization. Due to the capacity difference of the classifiers (one layer vs. up to three layers), there is a gap in the final classification accuracies of the two cases reported in Tables~\ref{tab:layer_ppi} and~\ref{tab:layer_taxi}. 
		}	
	\end{center}
\end{figure}

We report in Tables~\ref{tab:layer_ppi} and \ref{tab:layer_taxi} the classification accuracy when the classifier only consists of one layer and when the classifier is tuned using up to three layers (as described in Section.~\ref{subsubsec:hyperparameter}). 
In some cases, the 1-layer classifier could achieve an accuracy that is close or slightly higher than that of the classifier using up to three layers. However, tuning the model using up to three classification layers allows us to achieve better accuracy than using only one layer for the classifier in most cases. The few exceptions are highlighted in \red{red} in Tables~\ref{tab:layer_ppi} and \ref{tab:layer_taxi}. These results indicate that using a strong classifier is generally helpful in enhancing the classification accuracy. We also visualize in Fig.~\ref{fig:tsne2} the cases in which the differences between using one classification layer and using up to three classification layers are the highest. In particular, we highlight the t-SNE visualization of the embeddings generated by \textsc{GraphSage} \ourfinetune for the datasets \textsc{PTC-FM} and \textsc{JAN. G.} when using one classification layer and up to three classification layers, respectively.

\begin{table*}[t]	
	\caption{Classification accuracy when the classifier consists of one layer and when the classifier is tuned using up to three layers on the benchmark datasets. \ourtripletone and \ourtripletthree represent \ourtriplet with one classification layer and up to three classification layers (as detailed in Section.~\ref{subsubsec:hyperparameter}), respectively. The same applies for \ourfinetuneone and \ourfinetunethree. In almost all cases, tuning the classifier using up to three classifier layers leads to better accuracy than using only one layer.}
	\label{tab:layer_ppi}
	\centering
	\scalebox{0.83}{\renewcommand{\arraystretch}{1.0}
		{
		\begin{tabular}{@{}l|c||c|c|c|c|c|c@{}}
			\toprule
			\textbf{GNN} & Method  &\multicolumn{6}{c}{\textbf{Dataset}} \\ 
		    \cmidrule{3-8}
			& (\# Layers) & {\textsc{DD}} & {\textsc{MUTAG}} & {\textsc{MUTAG2}} & 
			{\textsc{PTC-FM}} &  {\textsc{PROTEINS}} & {\textsc{IMDB-B}}
			\\ 
			\midrule
				\textsc{GraphSage} & \ourtripletone & 74.92 $\pm$ 0.76 & 80.13 $\pm$ 1.07
		 & 75.63 $\pm$ 0.78	&  60.24 $\pm$ 1.13  & 69.87 $\pm$ 0.35 & \red{ 68.42 $\pm$ 0.45 } \\     
			& \ourtripletthree & 75.13 $\pm$ 0.82  & 80.86 $\pm$ 1.19
			  & 76.84 $\pm$ 0.54 & 62.75 $\pm$ 1.20 & 71.29 $\pm$ 0.41  &  68.37 $\pm$ 0.63 
			\\  
			\cmidrule{2-8}
			& \ourfinetuneone & 76.14 $\pm$ 1.33 & 80.97 $\pm$ 0.59 
			 & \red{77.82 $\pm$ 0.56} 
			&  60.13 $\pm$ 0.65  & 70.26 $\pm$ 0.53  & 68.17  $\pm$ 0.71 \\     
			& \ourfinetunethree & 76.52 $\pm$ 1.47  & 81.14 $\pm$ 0.68 
			  & 77.71 $\pm$ 0.41 &	62.65 $\pm$ 0.72  & 72.34 $\pm$ 0.56  & 68.24 $\pm$ 0.83 
			\\     
			\midrule
			\textsc{GAT}& \ourtripletone & 72.68 $\pm$ 0.84 & 76.13 $\pm$ 0.48 & 
			 75.32 $\pm$ 0.54 &  60.86 $\pm$ 1.04 & 69.73 $\pm$ 0.58 &   
			68.97 $\pm$ 0.82 
			 
			\\     
			 & \ourtripletthree & 72.95 $\pm$ 0.91 & 
			77.84 $\pm$ 0.63 &	76.34 $\pm$ 0.52  & 62.04 $\pm$ 1.16  & 
			70.17 $\pm$ 0.72 & 69.15 $\pm$ 0.87 
			\\
			\cmidrule{2-8}
			& \ourfinetuneone & 73.48 $\pm$ 1.29 & 78.02 $\pm$ 1.33 & 
			 75.25 $\pm$ 1.19 &  60.48 $\pm$ 0.72 & 70.67 $\pm$ 0.47 &   
			67.12 $\pm$ 0.65
			 
			\\     
			 & \ourfinetunethree & 74.13 $\pm$ 1.47 & 
			78.17 $\pm$ 1.41 & 
			76.49 $\pm$ 1.23  & 61.61 $\pm$ 0.53  &  
			72.64 $\pm$ 0.58 & 67.25 $\pm$ 0.89
			
			\\     
			\midrule
			\textsc{DiffPool} & \ourtripletone & 74.56 $\pm$ 0.48 & 83.69 $\pm$ 0.81
			 & 76.68 $\pm$ 1.05 & 60.93 $\pm$ 0.37  & 72.27 $\pm$ 0.48 &  
			64.93 $\pm$ 1.01 \\     
			& \ourtripletthree & 74.93 $\pm$ 0.53 & 86.14 $\pm$ 0.77
			 & 77.94 $\pm$ 1.28  & 62.03 $\pm$ 0.32  & 73.87 $\pm$ 0.64 & 
			 65.22	$\pm$ 0.83
			\\   
			\cmidrule{2-8}
			 & \ourfinetuneone & 77.03 $\pm$ 0.43 & 83.47 $\pm$ 0.59
			 & 76.37 $\pm$ 0.98 & 61.12 $\pm$ 0.47  &  72.18 $\pm$ 1.03 &  
			64.53 $\pm$ 0.69 \\     
			& \ourfinetunethree & 78.84 $\pm$ 0.54 & 87.38 $\pm$ 0.62
			 & 77.08 $\pm$ 1.23  & 62.15 $\pm$ 0.68  & 73.07 $\pm$ 1.17 & 64.90	$\pm$ 0.81
			\\   
			\midrule
			\textsc{EigenGCN} & \ourtripletone & 76.93 $\pm$ 0.37 & 80.15 $\pm$ 0.58
			 & 77.23 $\pm$ 0.95 &  63.87 $\pm$ 0.83 &  74.87 $\pm$ 0.34 &  
			70.45 $\pm$ 0.57 
			\\     
			 & \ourtripletthree & 77.56 $\pm$ 0.48 & 80.21 $\pm$ 0.71
			 & 77.98 $\pm$ 0.62  & 64.13 $\pm$ 0.95  & 75.93 $\pm$ 0.56 & 
			72.66	$\pm$ 0.42 \\
			\cmidrule{2-8}
			 & \ourfinetuneone & 76.88 $\pm$ 0.45 & 80.67 $\pm$ 0.71
			 & \red{77.12 $\pm$1.14}  & 63.41  $\pm$ 1.04 & 75.16 $\pm$ 1.37 &  
			71.13 $\pm$ 0.62 
			\\     
			 & \ourfinetunethree & 78.13 $\pm$ 0.51 & 81.42 $\pm$ 0.86
			 & 77.02 $\pm$ 1.72  & 63.52 $\pm$ 1.43  & 77.31 $\pm$ 1.46 & 
			72.04 $\pm$ 0.53
			\\     
			\midrule
			\textsc{SAGPool} & \ourtripletone & 77.15 $\pm$ 0.97 & 79.25 $\pm$ 0.87
			 & 76.73 $\pm$ 0.41 & 61.83 $\pm$ 0.75  & 76.57 $\pm$ 0.43 & 71.64 $\pm$ 0.83 
			\\     
			& \ourtripletthree & 78.32 $\pm$ 1.26 & 79.63 $\pm$ 0.95
			 & 78.03 $\pm$ 0.68   & 63.83 $\pm$ 0.83  & 77.52 $\pm$ 0.54 & 71.73 $\pm$ 0.81 
			\\ 
			\cmidrule{2-8}
			 & \ourfinetuneone & 77.84 $\pm$ 0.85 & 78.65 $\pm$ 0.91
			 & 76.79 $\pm$ 0.55 & 62.14 $\pm$ 0.77  & 75.32 $\pm$ 0.88 &  
			 71.84 $\pm$ 0.67 
			\\     
			& \ourfinetunethree & 78.22 $\pm$ 0.70 & 79.03 $\pm$ 0.89
			 & 77.03 $\pm$ 0.63   & 64.34 $\pm$ 0.86  & 76.23 $\pm$ 1.12 & 
			72.36 $\pm$ 0.73 
			\\ 
			\bottomrule
	\end{tabular}
	}
}
\end{table*}

\begin{table*}[htbp!]
	\caption{Classification accuracy when the classifier consists of one layer and when the classifier is tuned using up to three layers on the New York City Taxi datasets. \ourtripletone and \ourtripletthree represent \ourtriplet with one classification layer and up to three classification layers (as detailed in Section.~\ref{subsubsec:hyperparameter}), respectively. The same applies for \ourfinetuneone and \ourfinetunethree. Except for few cases, tuning the model by using up to three classification layers helps achieve better accuracy than using only one layer.}
	\label{tab:layer_taxi}
	\centering
	\scalebox{0.83}{ \renewcommand{\arraystretch}{1.0}
		{
		\begin{tabular}{@{}l|c||c|c|c|c|c|c@{}}
			\toprule
			\textbf{GNN} & Method  &\multicolumn{6}{c}{\textbf{Dataset}} \\ 
		    \cmidrule{3-8}
			& (\# Layers) &{\textsc{Jan. G.}} & {\textsc{Feb. G.}} & {\textsc{Mar. 
			G.}} & {\textsc{Jan. Y.}} & {\textsc{Feb. Y.}} & 
			{\textsc{Mar. Y.}} 
			\\ \midrule
			\textsc{GraphSage} & \ourtripletone & 74.81 $\pm$ 1.04 & 66.31 $\pm$ 1.05
		 & 67.04 $\pm$ 0.77 
			&  75.19 $\pm$ 1.25  & \red{65.51 $\pm$ 0.67}  & 70.04  $\pm$ 0.87 \\     
			& \ourtripletthree & 76.14 $\pm$ 0.93  & 66.67 $\pm$ 1.31
			  & 67.13 $\pm$ 0.85 & 75.24 $\pm$ 1.16 & 65.43 $\pm$ 0.68  & 70.15  $\pm$ 0.64 
			\\  
			\cmidrule{2-8}
			& \ourfinetuneone & 75.26 $\pm$ 0.77 & 66.59 $\pm$ 0.75
			 & 67.83 $\pm$ 0.98 
			&  \red{75.43 $\pm$ 1.38}  & 66.42 $\pm$ 0.46  &  69.83 $\pm$ 1.17 \\     
			& \ourfinetunethree & 76.63 $\pm$ 0.82  & 67.74 $\pm$ 0.88
			  & 68.95 $\pm$ 1.41 & 75.21 $\pm$ 1.70  & 67.64 $\pm$ 0.73  & 70.23  $\pm$ 1.25 
			\\     
			\midrule
			\textsc{GAT}& \ourtripletone & 75.07 $\pm$ 0.86 & \red{67.35 $\pm$ 0.73} & 67.15		 $\pm$ 0.89 &  75.27 $\pm$ 1.19 & 66.12 $\pm$ 0.74 &   
			70.23 $\pm$ 0.65
			 
			\\     
			 & \ourtripletthree & 75.23 $\pm$ 0.82 & 67.24 $\pm$ 0.56 & 
			 67.34 $\pm$ 0.71 & 76.82 $\pm$ 1.23  &  
			66.45 $\pm$ 0.85 & 70.66 $\pm$ 0.78
			 \\
			\cmidrule{2-8}
			& \ourfinetuneone  & 74.38 $\pm$ 0.91 &  67.74$\pm$ 0.65 & 
			68.26 $\pm$ 0.95 &  74.71 $\pm$ 1.14 & 66.57 $\pm$ 0.71 & \red{70.47 $\pm$ 0.98} 
			 
			\\     
			 & \ourfinetunethree & 74.65 $\pm$ 0.98 & 68.11 $\pm$ 0.69 & 69.15 $\pm$ 1.37
			 & 74.79 $\pm$ 1.27  & 68.75 $\pm$ 0.66 &  70.44	$\pm$ 0.93
			
			\\     
			\midrule
			\textsc{DiffPool} & \ourtripletone & 79.44 $\pm$ 1.07 & 73.18 $\pm$ 1.13
			 & 71.24 $\pm$ 0.77 & 74.37 $\pm$ 0.68  & 67.43 $\pm$ 0.73 &71.94 $\pm$ 0.61 \\     
			& \ourtripletthree & 80.28 $\pm$ 1.16 & 75.69 $\pm$ 1.21
			 & 73.79 $\pm$ 0.81  & 75.09 $\pm$ 0.72  & 68.19 $\pm$ 0.50 & 74.87	$\pm$ 0.83 
			\\   
			\cmidrule{2-8}
			 & \ourfinetuneone & 79.33 $\pm$ 0.95 & 73.29 $\pm$ 1.07
			 & 71.37 $\pm$ 0.59 & 74.88 $\pm$ 0.93  & 67.89 $\pm$ 0.54 &  
			 72.36 $\pm$ 0.75 \\     
			& \ourfinetunethree & 79.63 $\pm$ 0.82 & 74.56 $\pm$ 1.32
			 & 72.92 $\pm$ 0.65  & 75.96 $\pm$ 1.21  & 69.31 $\pm$ 0.97 &75.76	$\pm$ 0.86 
			\\   
			\midrule
			\textsc{EigenGCN} & \ourtripletone & 76.52 $\pm$ 0.97 & 68.71 $\pm$ 0.65
			 & 73.56 $\pm$ 1.45 & 73.47  $\pm$ 1.43 & 69.14 $\pm$ 0.85 & \red{70.06 $\pm$  0.81}
			\\     
			 & \ourtripletthree & 77.14 $\pm$ 0.81 & 70.03 $\pm$ 0.62
			 & 74.12 $\pm$ 1.34  & 74.36 $\pm$ 1.65  & 69.72 $\pm$ 0.97 &70.03	$\pm$ 0.86 \\
			\cmidrule{2-8}
			 & \ourfinetuneone & \red{ 76.81 $\pm$ 1.13} & 69.53 $\pm$ 1.17
			 & \red{73.72 $\pm$ 1.51} &  74.78 $\pm$ 1/27 & 70.25 $\pm$ 1.35 &  
			71.29 $\pm$ 0.57 
			\\     
			 & \ourfinetunethree & 76.73 $\pm$ 1.21 & 71.27 $\pm$ 1.33
			 & 73.37 $\pm$ 1.85  & 75.33 $\pm$ 1.14  & 71.65 $\pm$ 1.67 & 71.84 $\pm$ 0.62
			\\     
			\midrule
			\textsc{SAGPool} & \ourtripletone &  75.67$\pm$ 1.14 & 68.23 $\pm$ 1.17
			 & 73.58 $\pm$ 1.28 & 71.03 $\pm$ 0.67  & 69.37 $\pm$ 0.54 &  
			69.64 $\pm$ 0.97 
			\\     
			& \ourtripletthree & 76.36 $\pm$ 1.37 &  69.07$\pm$ 1.48 &  74.34$\pm$1.52& 71.11 $\pm$  0.73 & 70.02 $\pm$ 0.64 & 70.04 $\pm$  1.48
			\\ 
			\cmidrule{2-8}
			 & \ourfinetuneone & 74.89 $\pm$ 1.03 & 68.93 $\pm$ 0.99
			 & 73.06 $\pm$ 1.09 & 72.43 $\pm$ 0.65  & 69.05 $\pm$ 0.73 &  
			 69.83 $\pm$ 0.53 
			\\     
			& \ourfinetunethree & 75.38 $\pm$ 0.86 & 69.27 $\pm$ 1.12
			 & 73.19 $\pm$ 1.34   & 72.51 $\pm$ 0.85  & 69.16 $\pm$ 0.69 & 
			70.59 $\pm$ 0.52  
			\\ 
			\bottomrule
	\end{tabular}
	}
}
\end{table*}

%
%


\bibliographystyle{plain}
\bibliography{paper.bib}

\begin{thebibliography}{10}

\bibitem{chdchik2010large}
Gal Chechik, Varun Sharma, Uri Shalit, and Samy Bengio.
\newblock Large scale online learning of image similarity through ranking.
\newblock {\em Journal of Machine Learning Research}, 11(3), 2010.

\bibitem{chen2020measuring}
Deli Chen, Yankai Lin, Wei Li, Peng Li, Jie Zhou, and Xu~Sun.
\newblock Measuring and relieving the over-smoothing problem for graph neural
  networks from the topological view.
\newblock In {\em Proceedings of the AAAI Conference on Artificial
  Intelligence}, 2020.

\bibitem{ching2018opportunities}
Travers Ching, Daniel~S Himmelstein, Brett~K Beaulieu-Jones, Alexandr~A
  Kalinin, Brian~T Do, Gregory~P Way, Enrico Ferrero, Paul-Michael Agapow,
  Michael Zietz, Michael~M Hoffman, et~al.
\newblock Opportunities and obstacles for deep learning in biology and
  medicine.
\newblock {\em Journal of The Royal Society Interface}, 15(141):20170387, 2018.

\bibitem{dai2016discriminative}
Hanjun Dai, Bo~Dai, and Le~Song.
\newblock Discriminative embeddings of latent variable models for structured
  data.
\newblock In {\em Proceedings of the International Conference on Machine
  Learning}, 2016.

\bibitem{defferrard2016convolutional}
Micha{\"e}l Defferrard, Xavier Bresson, and Pierre Vandergheynst.
\newblock Convolutional neural networks on graphs with fast localized spectral
  filtering.
\newblock In {\em Proceedings of the International Conference on Neural
  Information Processing Systems}, 2016.

\bibitem{duvenaud2015convolutional}
David Duvenaud, Dougal Maclaurin, Jorge Aguilera-Iparraguirre, Rafael
  G{\'o}mez-Bombarelli, Timothy Hirzel, Al{\'a}n Aspuru-Guzik, and Ryan~P
  Adams.
\newblock Convolutional networks on graphs for learning molecular fingerprints.
\newblock In {\em Proceedings of the International Conference on Neural
  Information Processing Systems}, 2015.

\bibitem{gao2019graph}
Hongyang Gao and Shuiwang Ji.
\newblock Graph u-nets.
\newblock In {\em Proceedings of the International Conference on Machine
  Learning}, 2019.

\bibitem{hajebi2011fast}
Kiana Hajebi, Yasin Abbasi-Yadkori, Hossein Shahbazi, and Hong Zhang.
\newblock Fast approximate nearest-neighbor search with k-nearest neighbor
  graph.
\newblock In {\em Proceedings of the International Joint Conference on
  Artificial Intelligence}, 2011.

\bibitem{hamiltoninductive}
William~L Hamilton, Rex Ying, and Jure Leskovec.
\newblock Inductive representation learning on large graphs.
\newblock In {\em Proceedings of the International Conference on Neural
  Information Processing Systems}, 2017.

\bibitem{hlugzlplstrategies}
Weihua Hu, Bowen Liu, Joseph Gomes, Marinka Zitnik, Percy Liang, Vijay Pande,
  and Jure Leskovec.
\newblock Strategies for pre-training graph neural networks.
\newblock In {\em Proceedings of the International Conference on Learning
  Representations}, 2020.

\bibitem{hwang2022ahp}
Hyunjin Hwang, Seungwoo Lee, Chanyoung Park, and Kijung Shin.
\newblock Ahp: Learning to negative sample for hyperedge prediction.
\newblock In {\em Proceedings of the International ACM SIGIR Conference on
  Research and Development in Information Retrieval}, 2022.

\bibitem{kipf2016semi}
Thomas~N Kipf and Max Welling.
\newblock Semi-supervised classification with graph convolutional networks.
\newblock In {\em Proceedings of the International Conference on Learning
  Represetations}, 2017.

\bibitem{ko2020monstor}
Jihoon Ko, Kyuhan Lee, Kijung Shin, and Noseong Park.
\newblock Monstor: an inductive approach for estimating and maximizing
  influence over unseen networks.
\newblock In {\em Proceedings of the IEEE/ACM International Conference on
  Advances in Social Networks Analysis and Mining}, 2020.

\bibitem{ktena2018metric}
Sofia~Ira Ktena, Sarah Parisot, Enzo Ferrante, Martin Rajchl, Matthew Lee, Ben
  Glocker, and Daniel Rueckert.
\newblock Metric learning with spectral graph convolutions on brain
  connectivity networks.
\newblock {\em NeuroImage}, 169:431--442, 2018.

\bibitem{lee2019self}
Junhyun Lee, Inyeop Lee, and Jaewoo Kang.
\newblock Self-attention graph pooling.
\newblock In {\em Proceedings of the International Conference on Machine
  Learning}, 2019.

\bibitem{ling2020hierarchical}
Xiang Ling, Lingfei Wu, Saizhuo Wang, Tengfei Ma, Fangli Xu, Alex~X Liu,
  Chunming Wu, and Shouling Ji.
\newblock Hierarchical graph matching networks for deep graph similarity
  learning.
\newblock {\em arXiv preprint arXiv:2007.04395}, 2020.

\bibitem{liu2019community}
Jiahao Liu, Guixiang Ma, Fei Jiang, Chun-Ta Lu, S~Yu Philip, and Ann~B Ragin.
\newblock Community-preserving graph convolutions for structural and functional
  joint embedding of brain networks.
\newblock In {\em Proceedings of the IEEE International Conference on Big
  Data}, 2019.

\bibitem{lu2021learning}
Yuanfu Lu, Xunqiang Jiang, Yuan Fang, and Chuan Shi.
\newblock Learning to pre-train graph neural networks.
\newblock In {\em Proceedings of the AAAI Conference on Artificial
  Intelligence}, 2021.

\bibitem{yao2019graph}
Yao Ma, Suhang Wang, Charu~C. Aggarwal, and Jiliang Tang.
\newblock Graph convolutional networks with eigenpooling.
\newblock In {\em Proceedings of the International Conference on Knowledge
  Discovery \& Data Mining}, 2019.

\bibitem{morris2020tudataset}
Christopher Morris, Nils~M Kriege, Franka Bause, Kristian Kersting, Petra
  Mutzel, and Marion Neumann.
\newblock Tudataset: A collection of benchmark datasets for learning with
  graphs.
\newblock {\em arXiv preprint arXiv:2007.08663}, 2020.

\bibitem{niepert2016learning}
Mathias Niepert, Mohamed Ahmed, and Konstantin Kutzkov.
\newblock Learning convolutional neural networks for graphs.
\newblock In {\em Proceedings of the International Conference on Machine
  Learning}, 2016.

\bibitem{park2015a}
Youngki Park, Heasoo Hwang, and Sang-goo Lee.
\newblock A fast k-nearest neighbor search using query-specific signature
  selection.
\newblock In {\em Proceedings of the ACM International on Conference on
  Information and Knowledge Management}, 2015.

\bibitem{qilearning}
Siyuan Qi, Wenguan Wang, Baoxiong Jia, Jianbing Shen, and Song-Chun Zhu.
\newblock Learning human-object interactions by graph parsing neural networks.
\newblock In {\em Proceedings of the European Conference on Computer Vision},
  2018.

\bibitem{schlichtkrull2018modeling}
Michael Schlichtkrull, Thomas~N Kipf, Peter Bloem, Rianne Van Den~Berg, Ivan
  Titov, and Max Welling.
\newblock Modeling relational data with graph convolutional networks.
\newblock In {\em Proceedings of the European Semantic Web Conference}, 2018.

\bibitem{schroff2015facenet}
Florian Schroff, Dmitry Kalenichenko, and James Philbin.
\newblock Facenet: A unified embedding for face recognition and clustering.
\newblock In {\em Proceedings of the IEEE Conference on Computer Vision and
  Pattern Recognition}, 2015.

\bibitem{shipoint}
Weijing Shi and Raj Rajkumar.
\newblock Point-gnn: Graph neural network for 3d object detection in a point
  cloud.
\newblock In {\em Proceedings of the IEEE/CVF conference on computer vision and
  pattern recognition}, 2020.

\bibitem{velickovic2018graph}
Petar Veli{\v{c}}kovi{\'{c}}, Guillem Cucurull, Arantxa Casanova, Adriana
  Romero, Pietro Li{\`{o}}, and Yoshua Bengio.
\newblock Graph attention networks.
\newblock In {\em Proceedings of the International Conference on Learning
  Representations}, 2018.

\bibitem{wang2019data}
Jingshu Wang, Divyansh Agarwal, Mo~Huang, Gang Hu, Zilu Zhou, Chengzhong Ye,
  and Nancy~R Zhang.
\newblock Data denoising with transfer learning in single-cell transcriptomics.
\newblock {\em Nature methods}, 16(9):875--878, 2019.

\bibitem{xiescale}
Guo-Sen Xie, Jie Liu, Huan Xiong, and Ling Shao.
\newblock Scale-aware graph neural network for few-shot semantic segmentation.
\newblock In {\em Proceedings of the IEEE/CVF Conference on Computer Vision and
  Pattern Recognition}, pages 5475--5484, 2021.

\bibitem{zhitao2018hierarchical}
Zhitao Ying, Jiaxuan You, Christopher Morris, Xiang Ren, Will Hamilton, and
  Jure Leskovec.
\newblock Hierarchical graph representation learning with differentiable
  pooling.
\newblock In {\em Proceedings of the International Conference on Neural
  Information Processing Systems}, 2018.

\bibitem{zhang2018end}
Muhan Zhang, Zhicheng Cui, Marion Neumann, and Yixin Chen.
\newblock An end-to-end deep learning architecture for graph classification.
\newblock In {\em Proceedings of the AAAI Conference on Artificial
  Intelligence}, 2018.

\end{thebibliography}

\end{document}